\documentclass[10pt,twocolumn,letterpaper]{article}

\usepackage[accsupp]{axessibility}
\usepackage{cvpr}              

\usepackage{times}
\usepackage{epsfig}
\usepackage{graphicx}
\usepackage{amsmath}
\usepackage{amssymb}
\usepackage{colortbl}
\usepackage{subcaption}
\usepackage[utf8]{inputenc} 
\usepackage[T1]{fontenc}    
\usepackage{url}            
\usepackage{booktabs}       
\usepackage{amsfonts}       
\usepackage{nicefrac}       
\usepackage{microtype}      
\usepackage{xcolor, soul}         

\definecolor{Lavender}{RGB}{220,190,130}
\definecolor{myPaleGreen}{RGB}{110,190,110} 
\usepackage{pifont}

\usepackage{xcolor}    
\usepackage{overpic}   

\usepackage{mathtools} 
\usepackage{amsmath,amsfonts,amsthm,bm,mathtools, bbm}
\usepackage{amsthm}
\usepackage{amssymb}
\usepackage{eqparbox}
\usepackage{arydshln}
\usepackage{multirow}
\usepackage{fontawesome5}
\usepackage{verbatim}

\newcommand{\fig}[2][1]{\includegraphics[draft=False, width=#1\linewidth]{fig/#2}}

\definecolor{demphcolor}{gray}{.5}
\newcommand{\demph}[1]{\textcolor{demphcolor}{#1}}
\usepackage{tikz}
\usetikzlibrary{tikzmark}

\definecolor{gain}{RGB}{71, 67, 229} 
\definecolor{gain}{rgb}{0.21,0.49,0.74}
\newcommand{\gain}[1]{\textcolor{gain}{#1}}
\definecolor{lost}{RGB}{140, 64, 64} 

\newlength\savewidth\newcommand\shline{\noalign{\global\savewidth\arrayrulewidth
  \global\arrayrulewidth 1pt}\hline\noalign{\global\arrayrulewidth\savewidth}}
\newcommand{\res}[2]{{#1} {\small{({\gain{#2}})}}}

\definecolor{demphcolorinline}{gray}{.3}

\newcommand{\tablefirst}{\cellcolor{gray!10}}
\newcommand{\ours}{\textcolor{black}{LaSt-ViT}\xspace}
\newcommand{\ourmethod}{\textcolor{black}{LazyStrike}\xspace}

\usepackage{pifont}
\usepackage{listings}
\usepackage{upquote}  
\lstdefinestyle{overleaf}{
    backgroundcolor=\color[rgb]{0.95,0.95,0.92},   
    commentstyle=\color[rgb]{0,0.6,0},
    keywordstyle=\color{magenta},
    numberstyle=\tiny\color[rgb]{0.5,0.5,0.5},
    stringstyle=\color[rgb]{0.58,0,0.82},
    basicstyle=\ttfamily\footnotesize,
    breakatwhitespace=false,         
    breaklines=true,                 
    captionpos=b,                    
    keepspaces=true,                 
    numbers=left,                    
    numbersep=5pt,                  
    showspaces=false,                
    showstringspaces=false,
    showtabs=false,                  
    tabsize=2
}
\usepackage{algorithm}
\lstdefinestyle{mocov3}{
  backgroundcolor=\color{white},
  basicstyle=\fontsize{7.5pt}{7.5pt}\ttfamily\selectfont,
  columns=fullflexible,
  breaklines=true,
  captionpos=b,
  commentstyle=\fontsize{7.5pt}{7.5pt}\color[rgb]{0.25,0.5,0.5},
  keywordstyle=\fontsize{7.5pt}{7.5pt}\color[rgb]{0.85,0.18,0.50},
}
\usepackage{etoolbox}
\makeatletter
\AfterEndEnvironment{algorithm}{\let\@algcomment\relax}
\AtEndEnvironment{algorithm}{\kern2pt\hrule\relax\vskip3pt\@algcomment}
\let\@algcomment\relax
\newcommand\algcomment[1]{\def\@algcomment{\footnotesize#1}}
\renewcommand\fs@ruled{\def\@fs@cfont{\bfseries}\let\@fs@capt\floatc@ruled
  \def\@fs@pre{\hrule height.8pt depth0pt \kern2pt}%
  \def\@fs@post{}%
  \def\@fs@mid{\kern2pt\hrule\kern2pt}%
  \let\@fs@iftopcapt\iftrue}
\makeatother
\usepackage{adjustbox}

\usepackage{quoting}
\usepackage[most]{tcolorbox}
\newtcolorbox{conclude}[1][]{%
	breakable,
	top=6pt,
	bottom=6pt,
	left=6pt,
	right=6pt,
	boxrule=.6pt,
	sharp corners,
	colframe=black,
  #1
}

\makeatletter
\colorlet{qaframe}{black}
\colorlet{qaback}{white}

\newcommand{\setqapaircolor}[1]{%
  \colorlet{qaframe}{#1}%
  \colorlet{qaback}{#1!5}%
}

\definecolor{qaA}{HTML}{1F77B4} 
\definecolor{qaB}{HTML}{2CA02C} 
\definecolor{qaC}{HTML}{D62728} 
\definecolor{qaD}{HTML}{9467BD} 
\definecolor{qaE}{HTML}{FF7F0E} 

\newcommand{\useqaA}{\setqapaircolor{qaA}}
\newcommand{\useqaB}{\setqapaircolor{qaB}}

\makeatother

\newtcolorbox{question}[1][]{%
  breakable,
  top=1pt, bottom=1pt, left=6pt, right=6pt,
  boxrule=.6pt,
  sharp corners,
  colframe=qaframe,
  colback=qaback,
  before upper={\textbf{Q: }\ignorespaces}, 
  #1
}

\newtcolorbox{answer}[1][]{%
  breakable,
  top=1pt, bottom=1pt, left=6pt, right=6pt,
  boxrule=.6pt,
  sharp corners,
  colframe=qaframe,
  colback=qaback,
  before upper={\textbf{A: }\ignorespaces}, 
  #1
}

\definecolor{gray90}{gray}{.90}

\makeatletter
\renewcommand\@fnsymbol[1]{%
  \ensuremath{%
    \ifcase#1\or \dagger\or \ddagger\or \mathsection\or \mathparagraph\or
    \|\or *\or \dagger\dagger\or \ddagger\ddagger \else\@ctrerr\fi
  }%
}
\makeatother

\definecolor{cvprblue}{rgb}{0.21,0.49,0.74}
\usepackage[pagebackref,breaklinks,colorlinks,allcolors=cvprblue]{hyperref}

\def\paperID{\#\#\#\#} 
\def\confName{CVPR}
\def\confYear{2026}

\title{Vision Transformers Need More Than Registers}

\author{Cheng Shi$^{1}$\hspace{1.0em} Yizhou Yu$^{1}$$^{\dagger}$\hspace{1.0em} Sibei Yang$^{2}$$^{\dagger}$ \\
 $^{1}$School of Computing and Data Science, The University of Hong Kong\hspace{1.0em} $^{2}$Sun Yat-sen University \\
{\footnotesize\texttt{shicheng2025@connect.hku.hk},\hspace{1.0em}
\texttt{yizhouy@acm.org}, \hspace{1.0em}
\texttt{yangsb3@mail.sysu.edu.cn}} \\
{\footnotesize \url{https://github.com/ChengShiest/LAST-ViT}}
 }

\begin{document}

\twocolumn[{
\maketitle
\vspace{-35pt}
\begin{center}
    \centering
\vspace{3mm}
\small
\captionsetup{type=figure}
\begin{tabular}{c@{\hspace{8pt}} c@{\hspace{0.8pt}}c@{\hspace{0.8pt}}c@{\hspace{8pt}}c@{\hspace{0.8pt}}c@{\hspace{0.8pt}}c@{\hspace{8pt}}c@{\hspace{0.8pt}}c@{\hspace{0.8pt}}c}
    input                    & \multicolumn{3}{c}{ConvNet~\cite{resnet,dinov1}} & \multicolumn{3}{c}{Transformer~\cite{vit,dinov1}} & \multicolumn{3}{c}{Transformer w/ \ourmethod (Ours)} \\
    \fig[0.09]{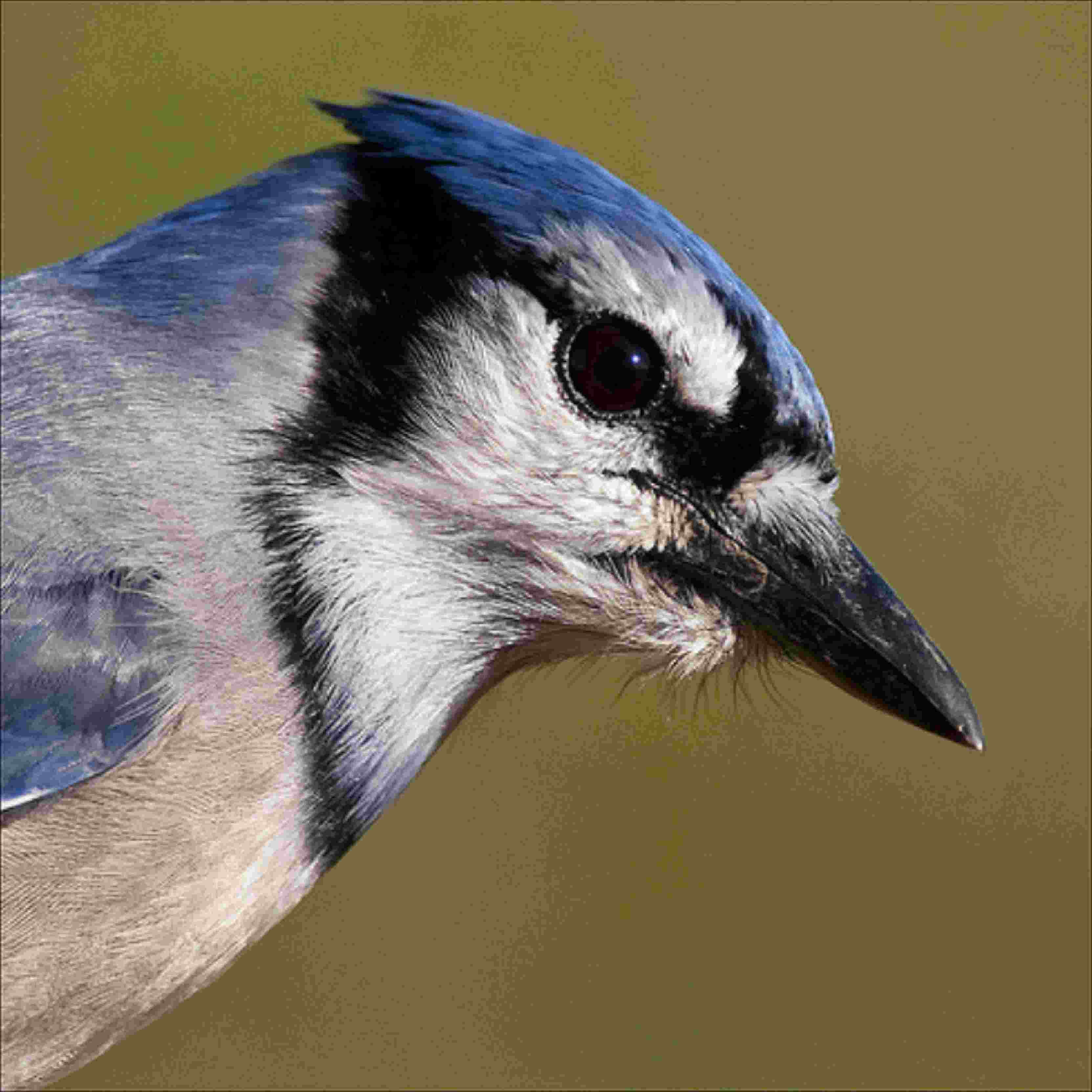} & \fig[0.09]{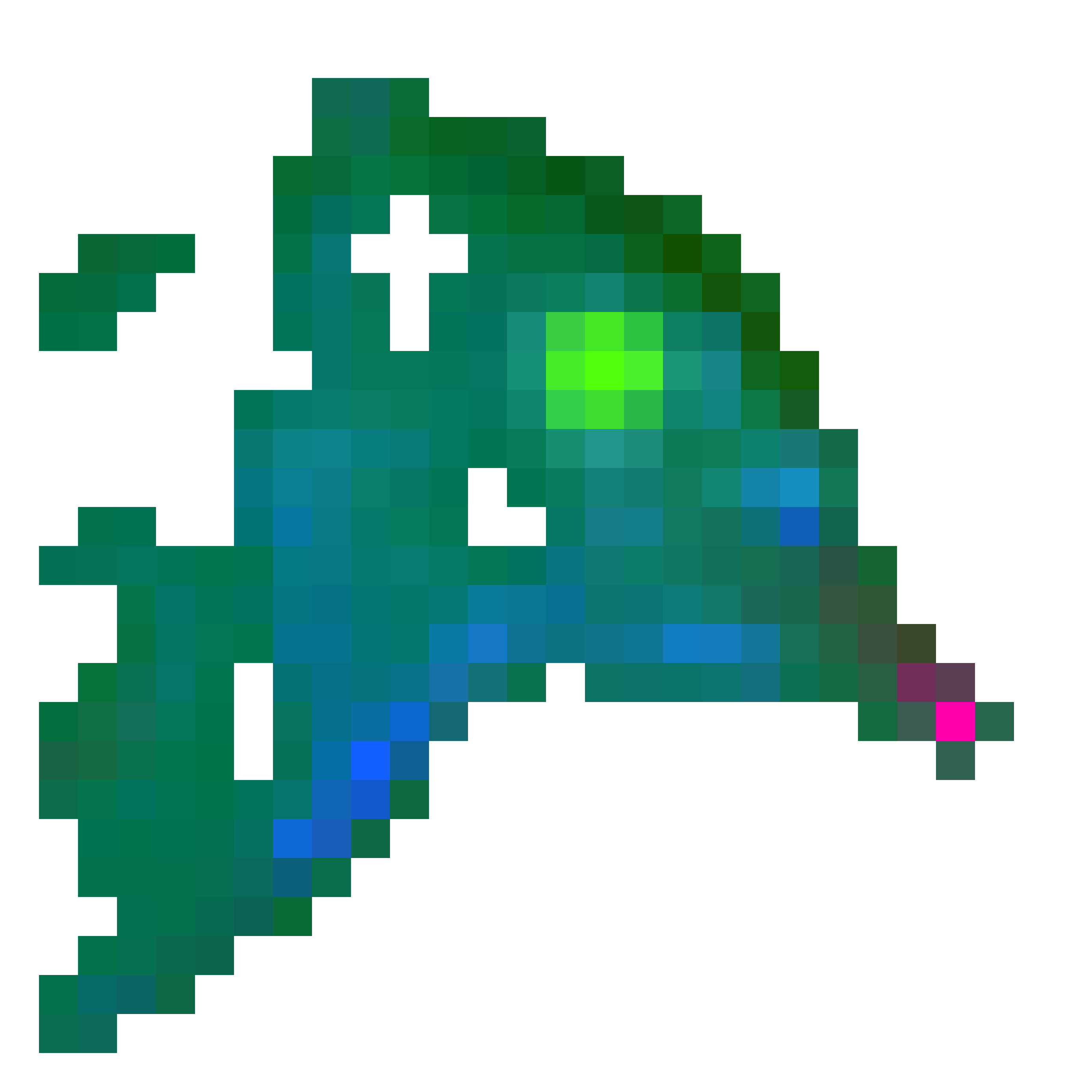}                  & \fig[0.09]{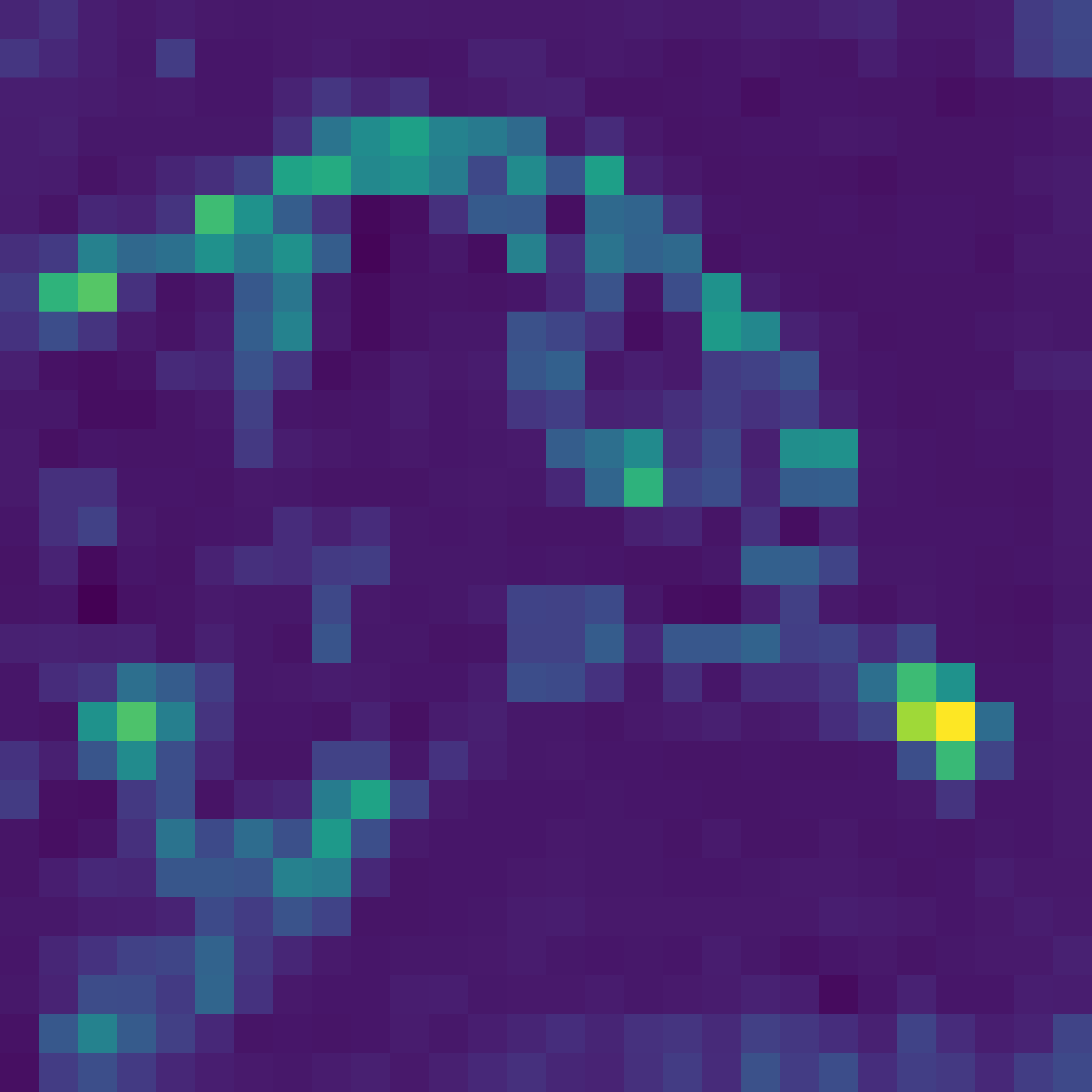}                                       & \fig[0.09]{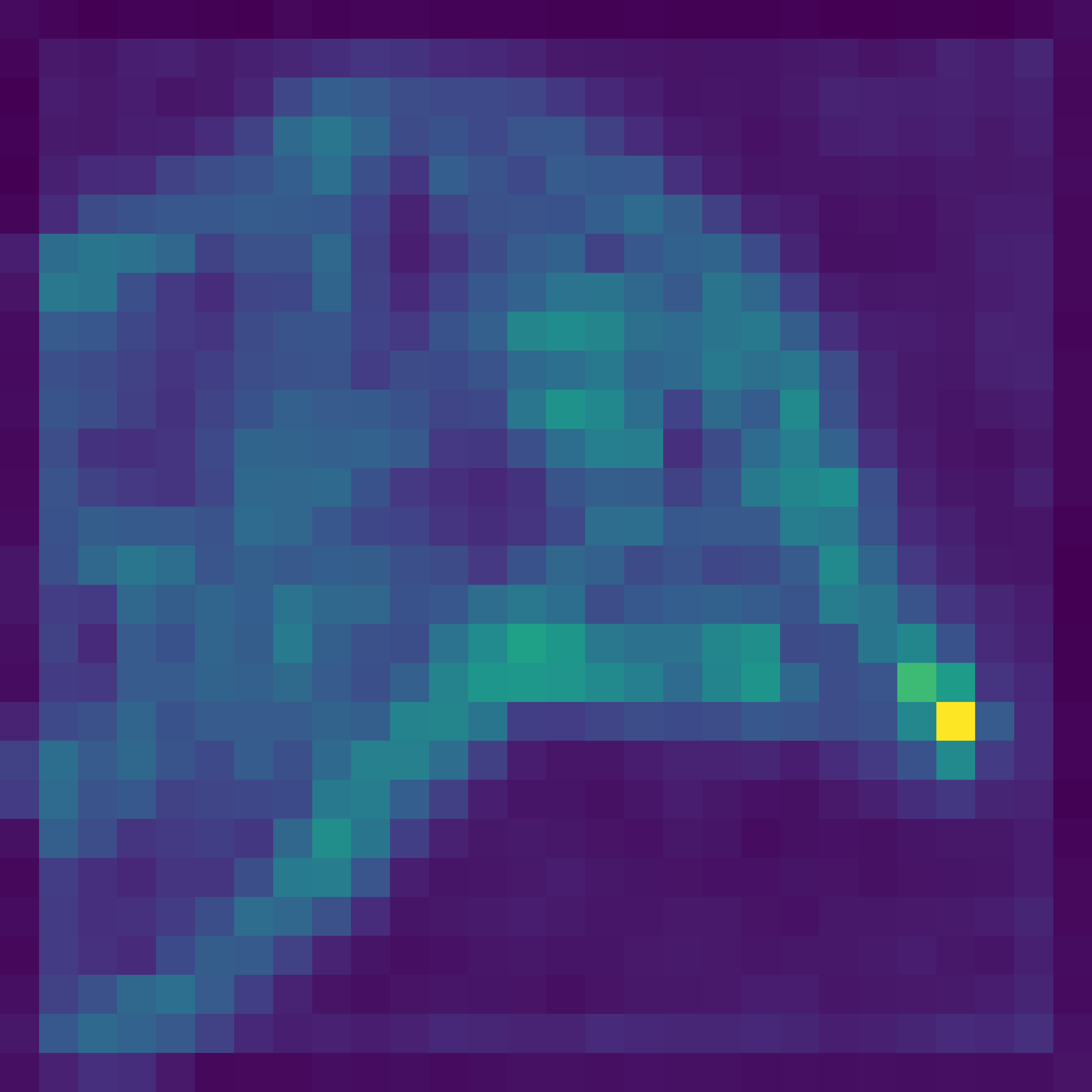}                          & \fig[0.09]{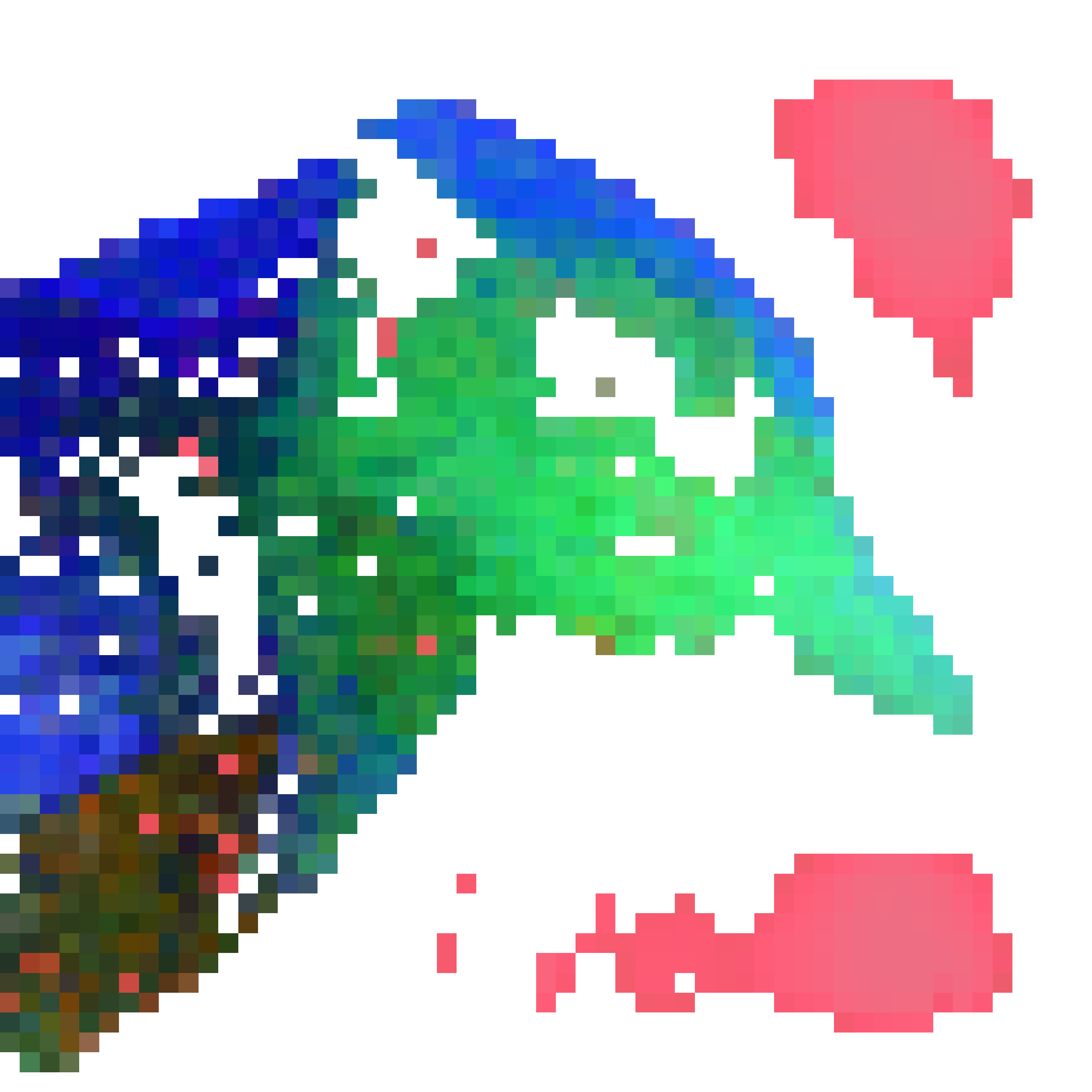} & \fig[0.09]{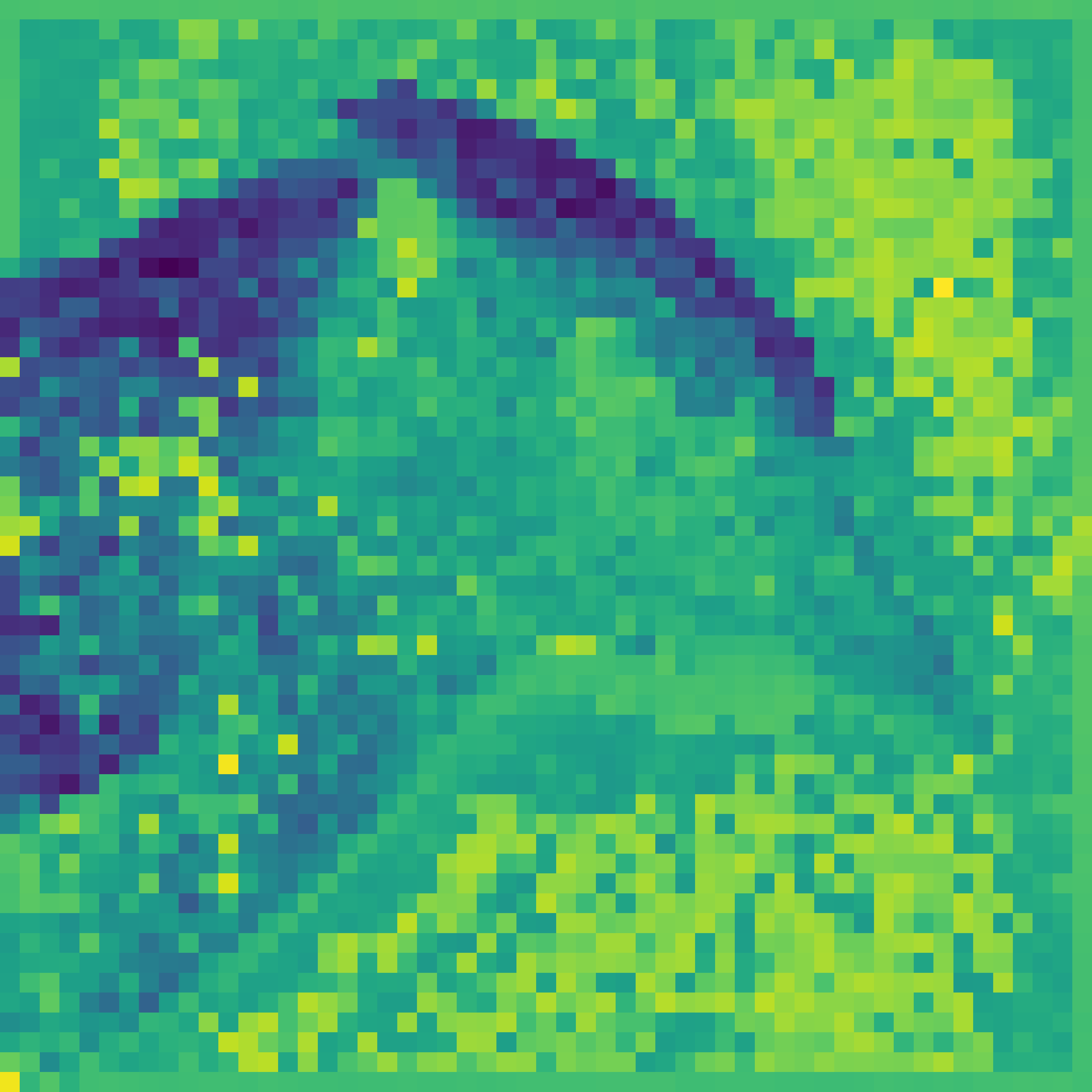} & \fig[0.09]{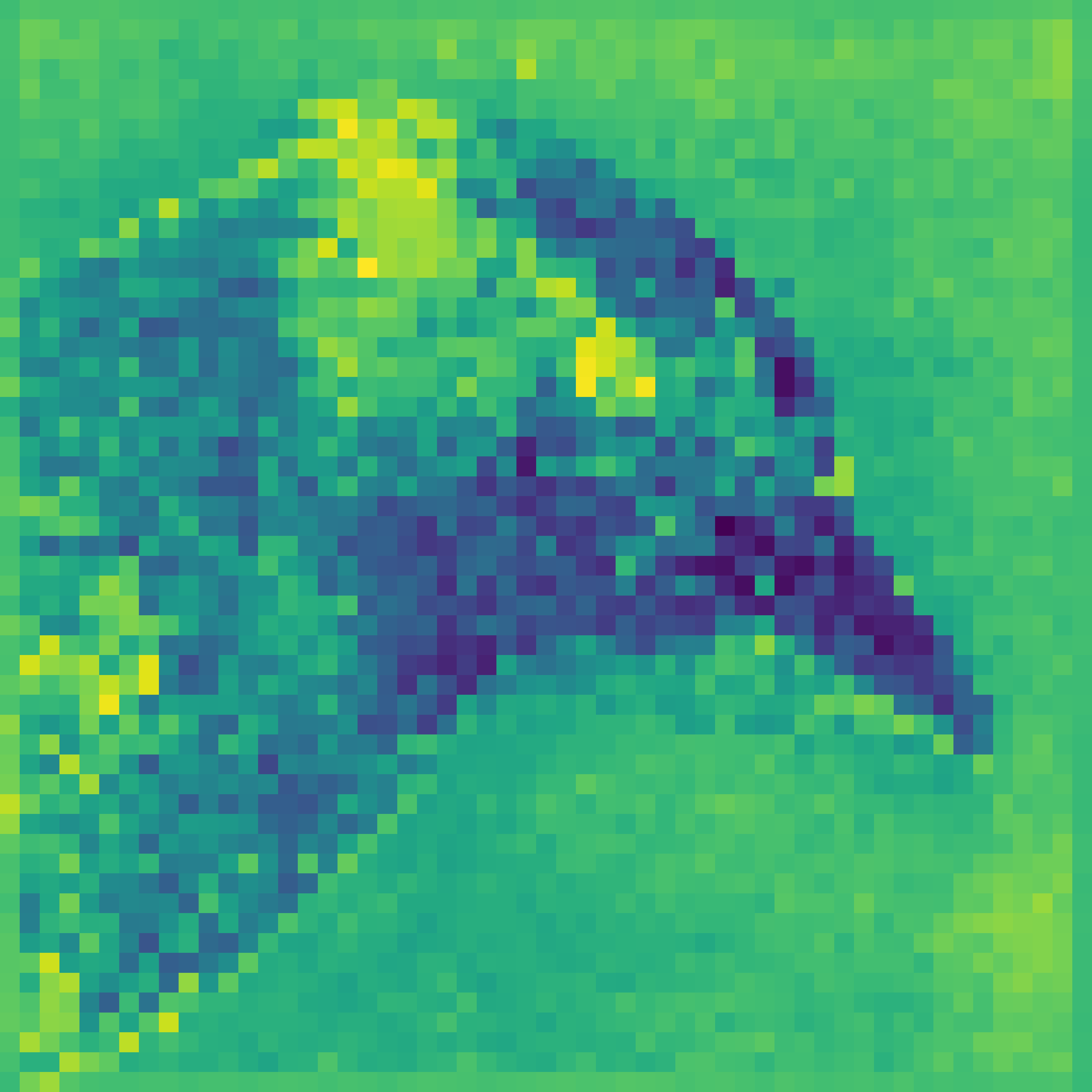} & \fig[0.09]{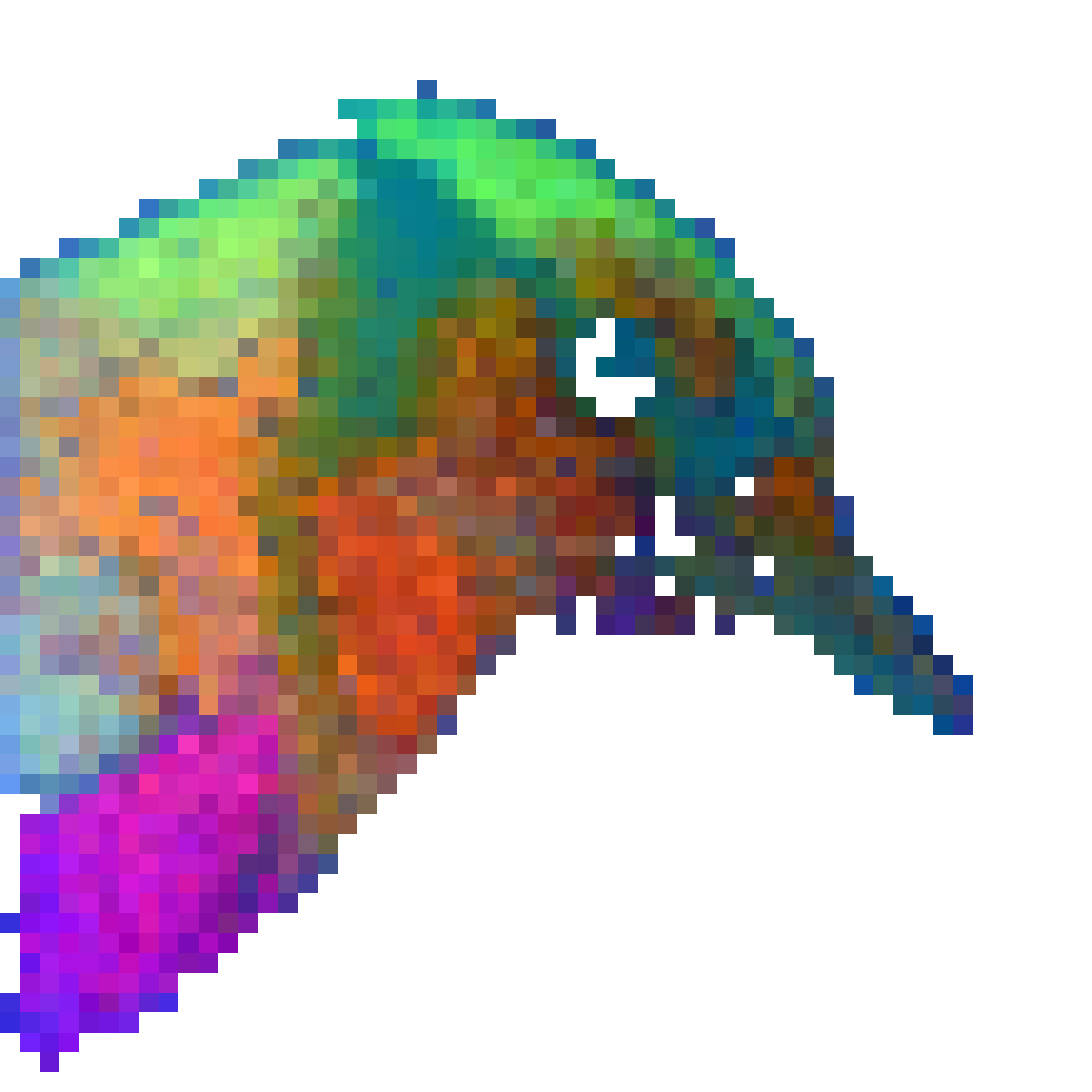} & \fig[0.09]{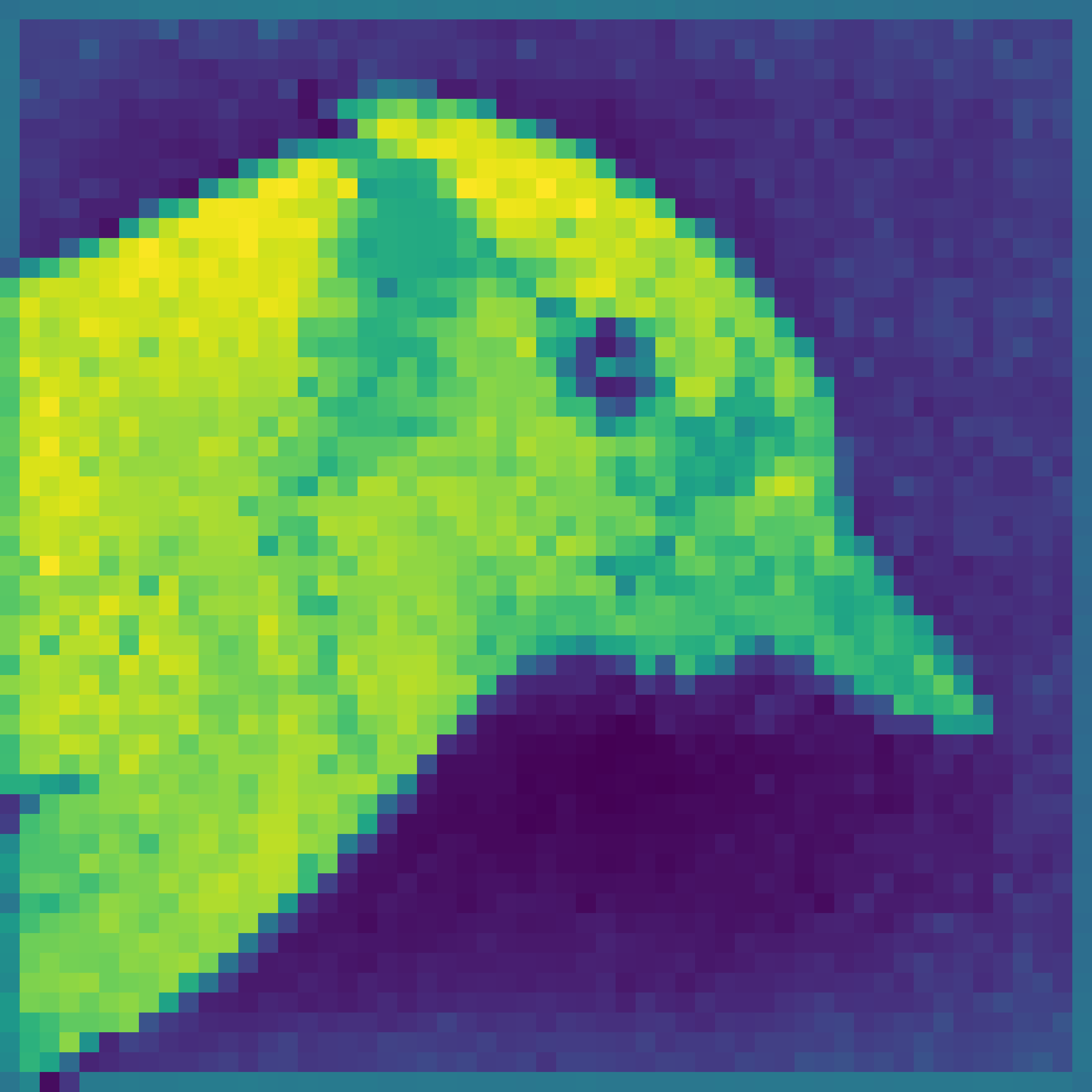} & \fig[0.09]{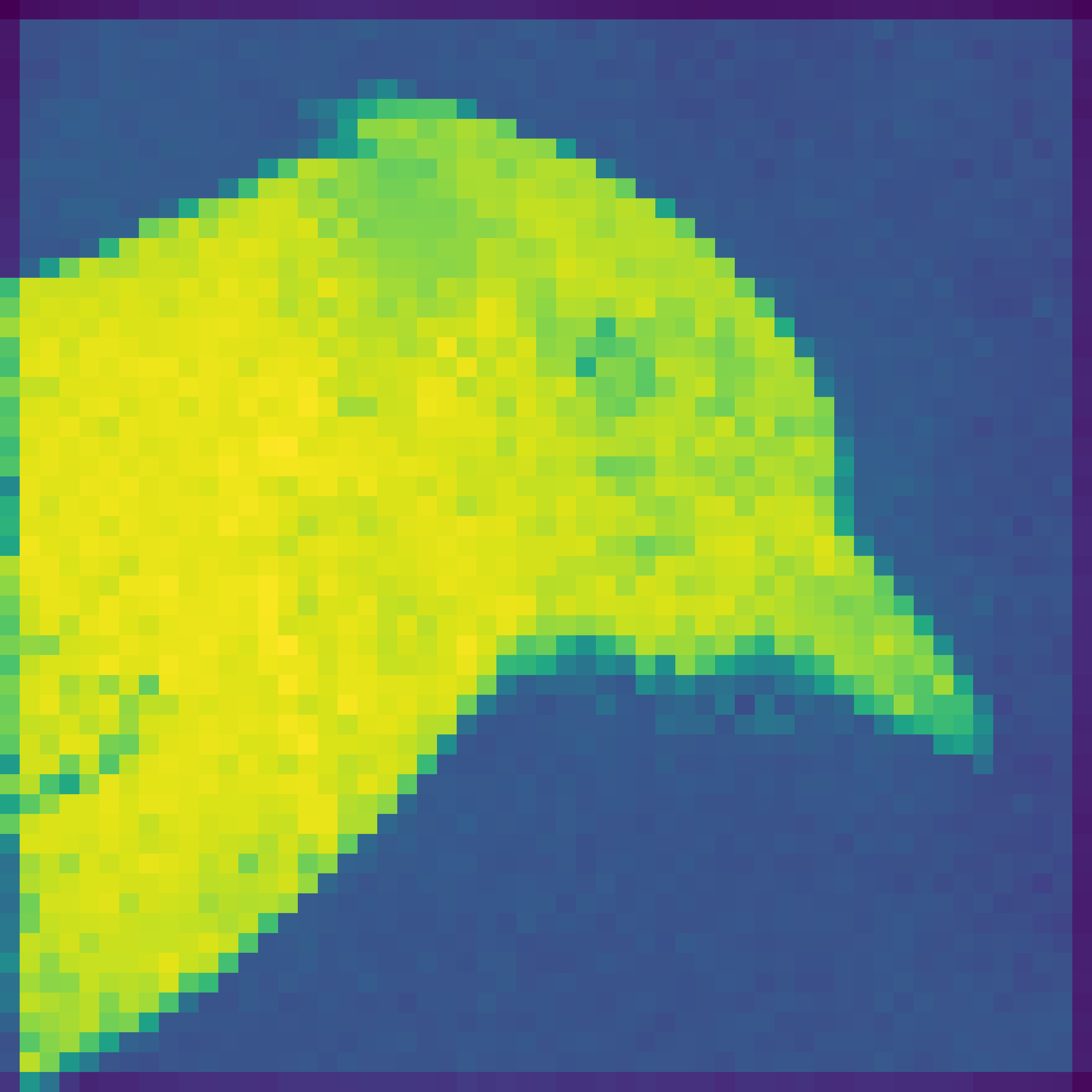} \\
    \fig[0.09]{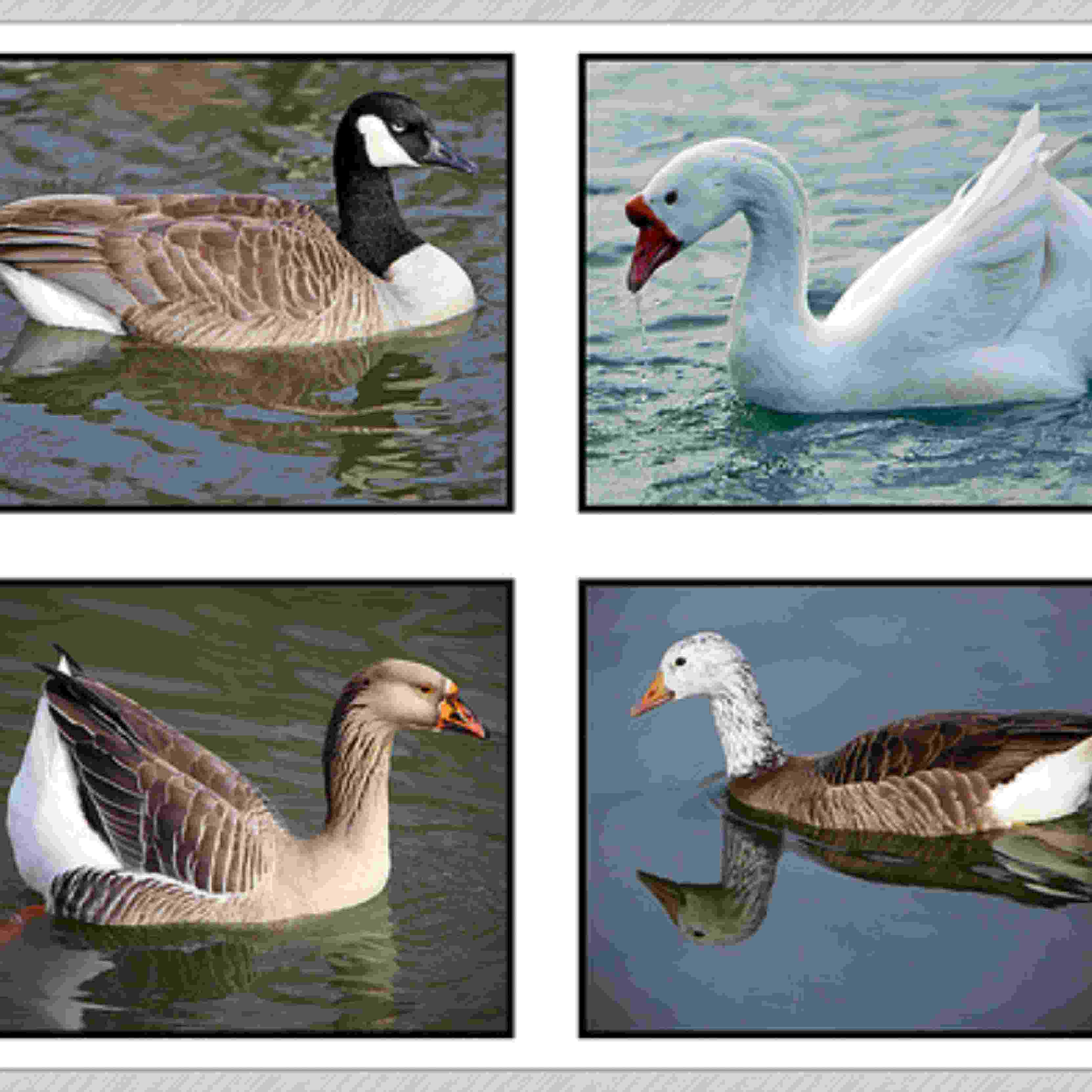} & \fig[0.09]{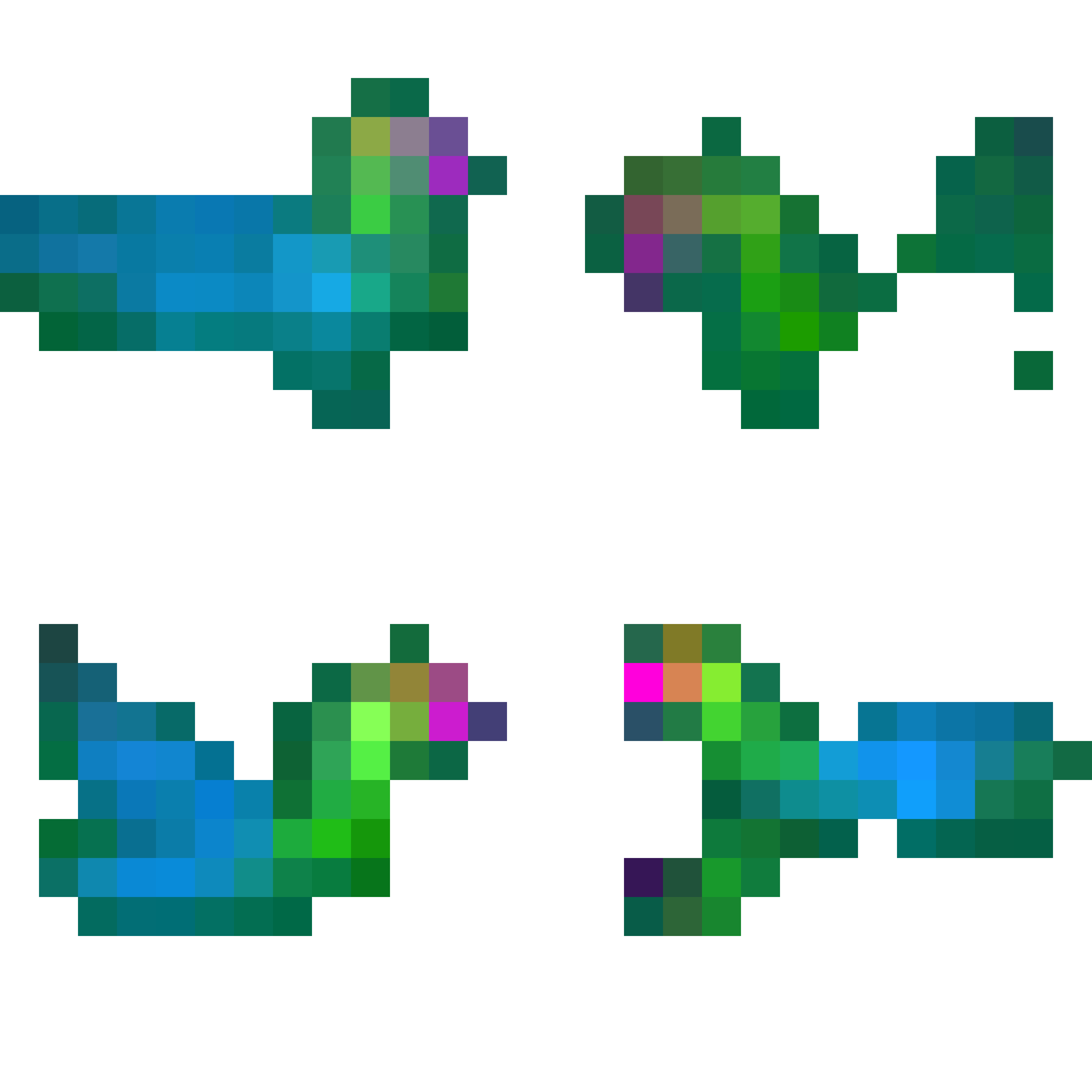}                  & \fig[0.09]{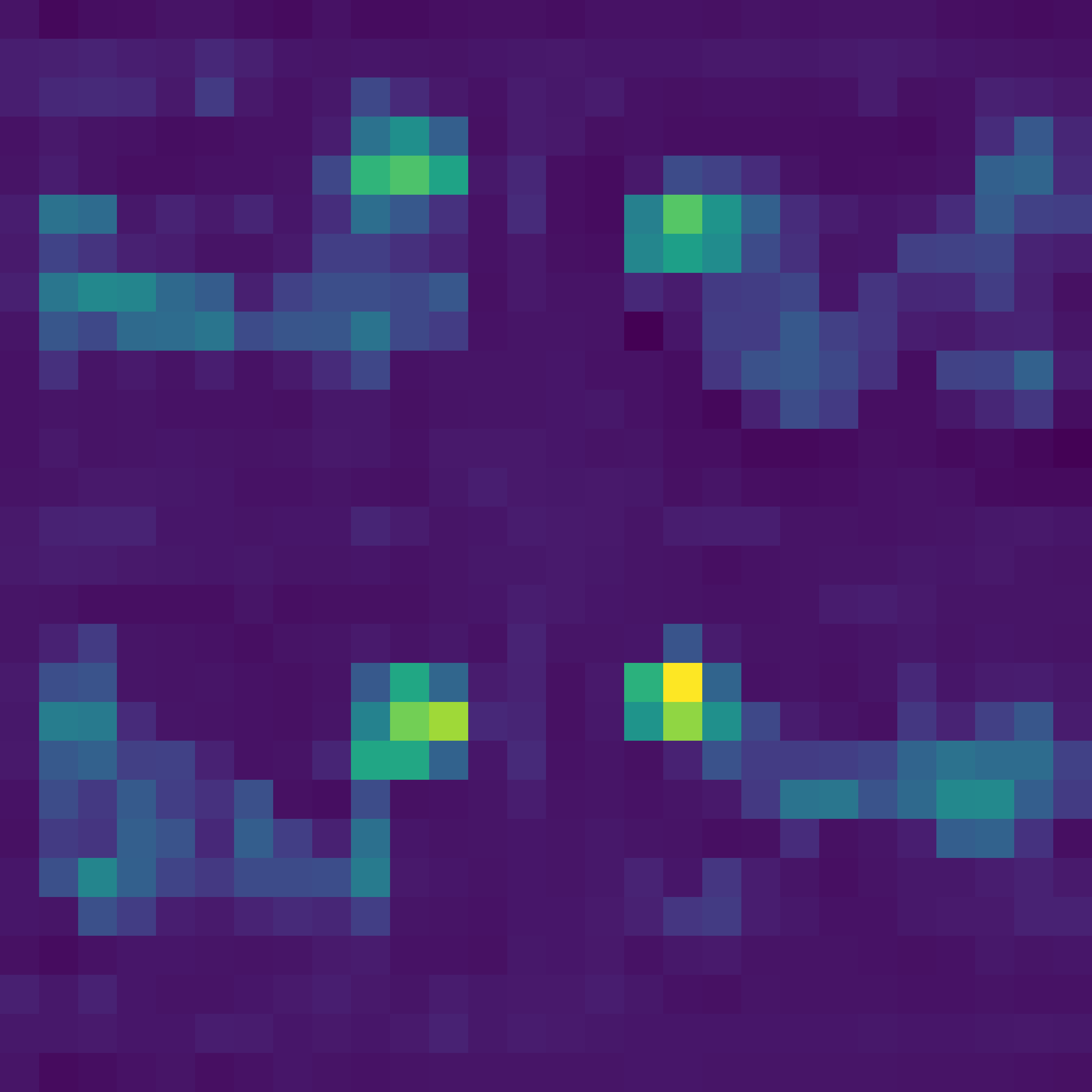}                                       & \fig[0.09]{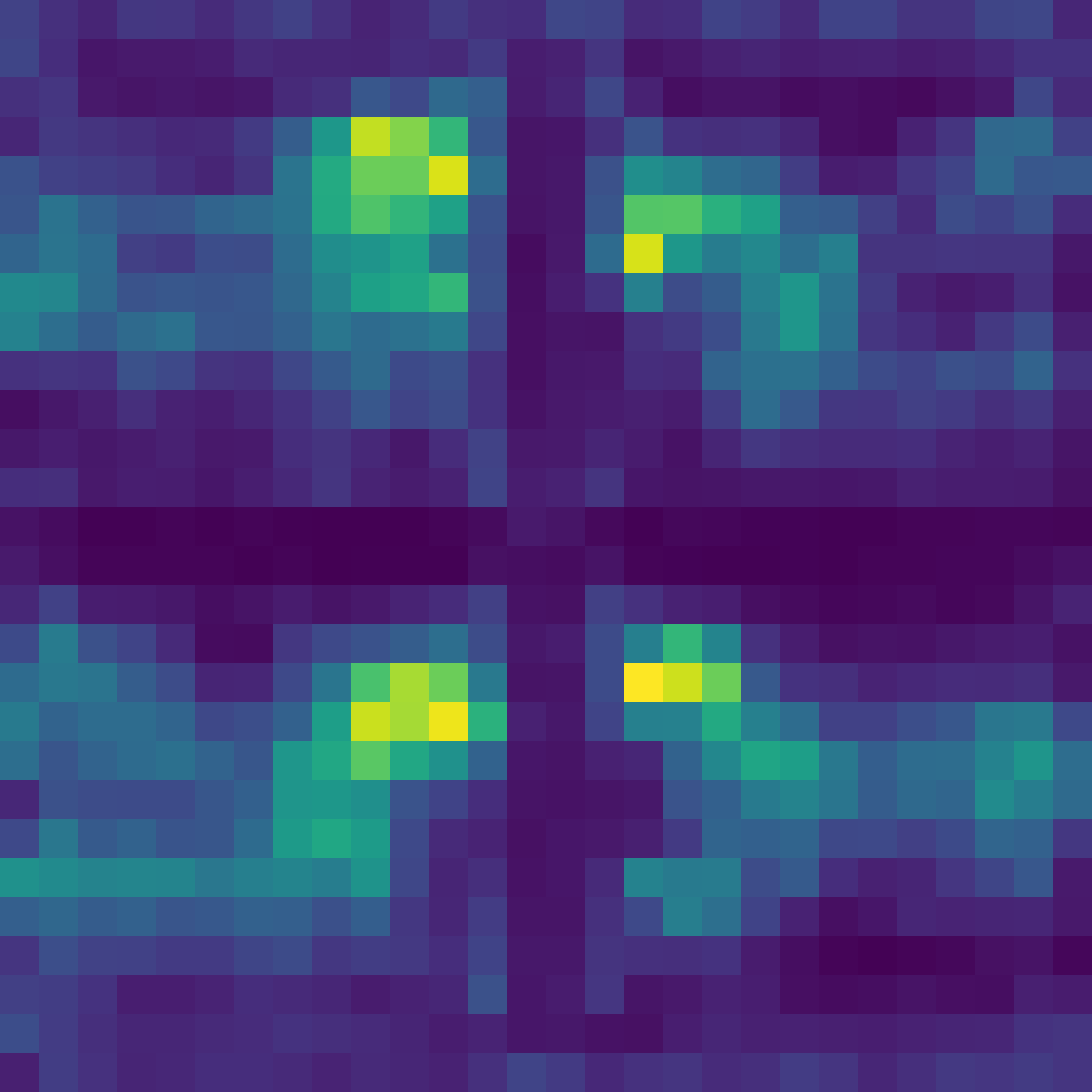}                          & \fig[0.09]{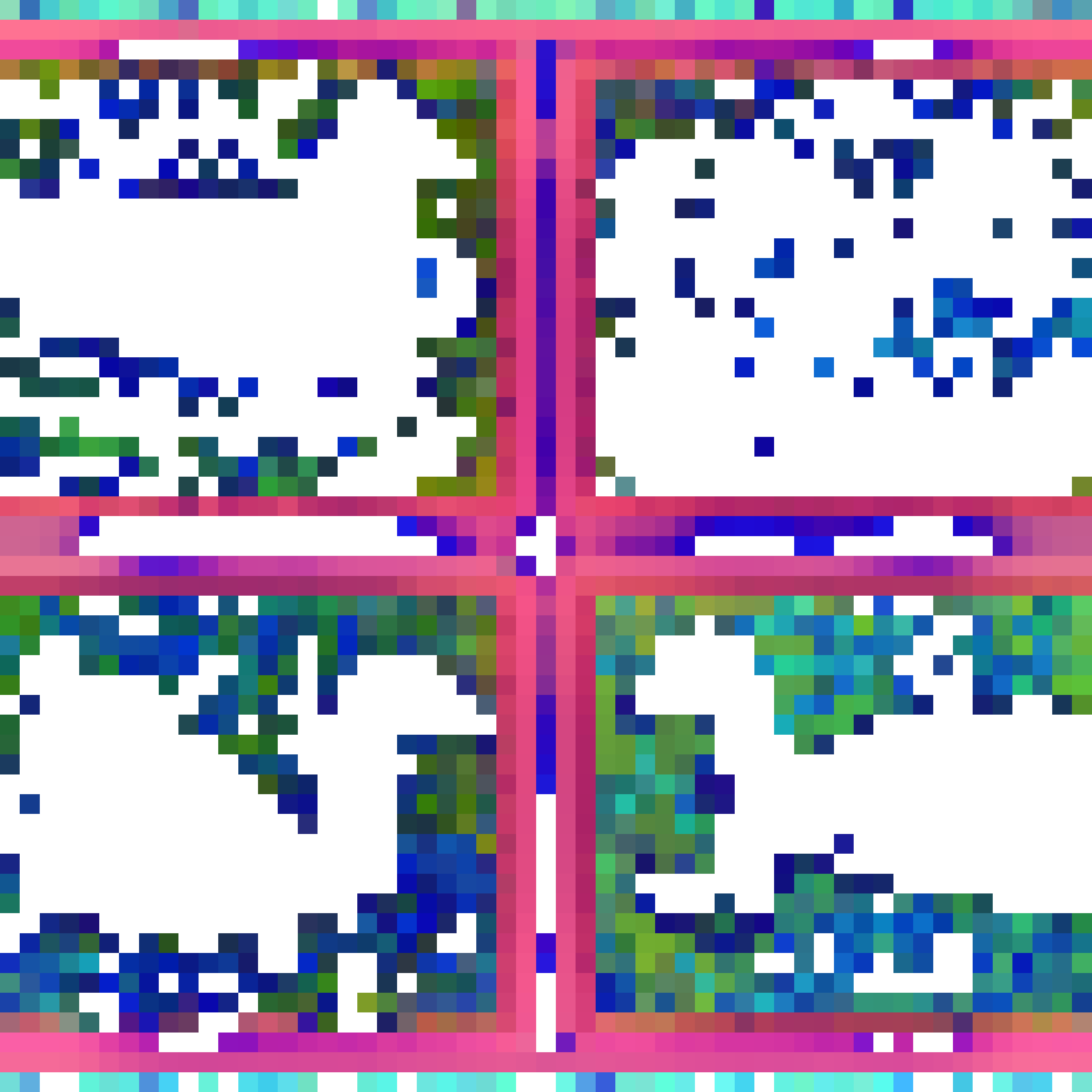} & \fig[0.09]{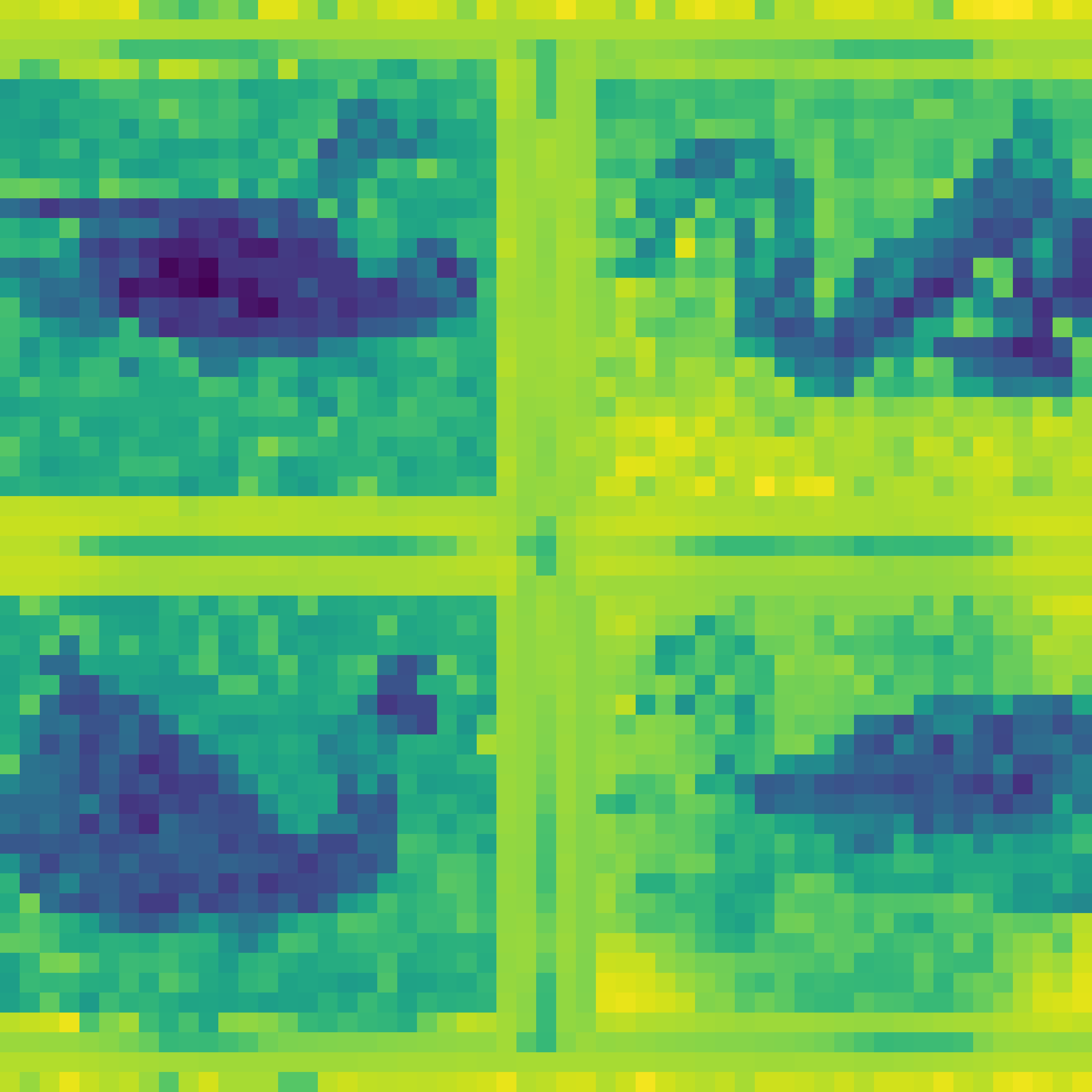} & \fig[0.09]{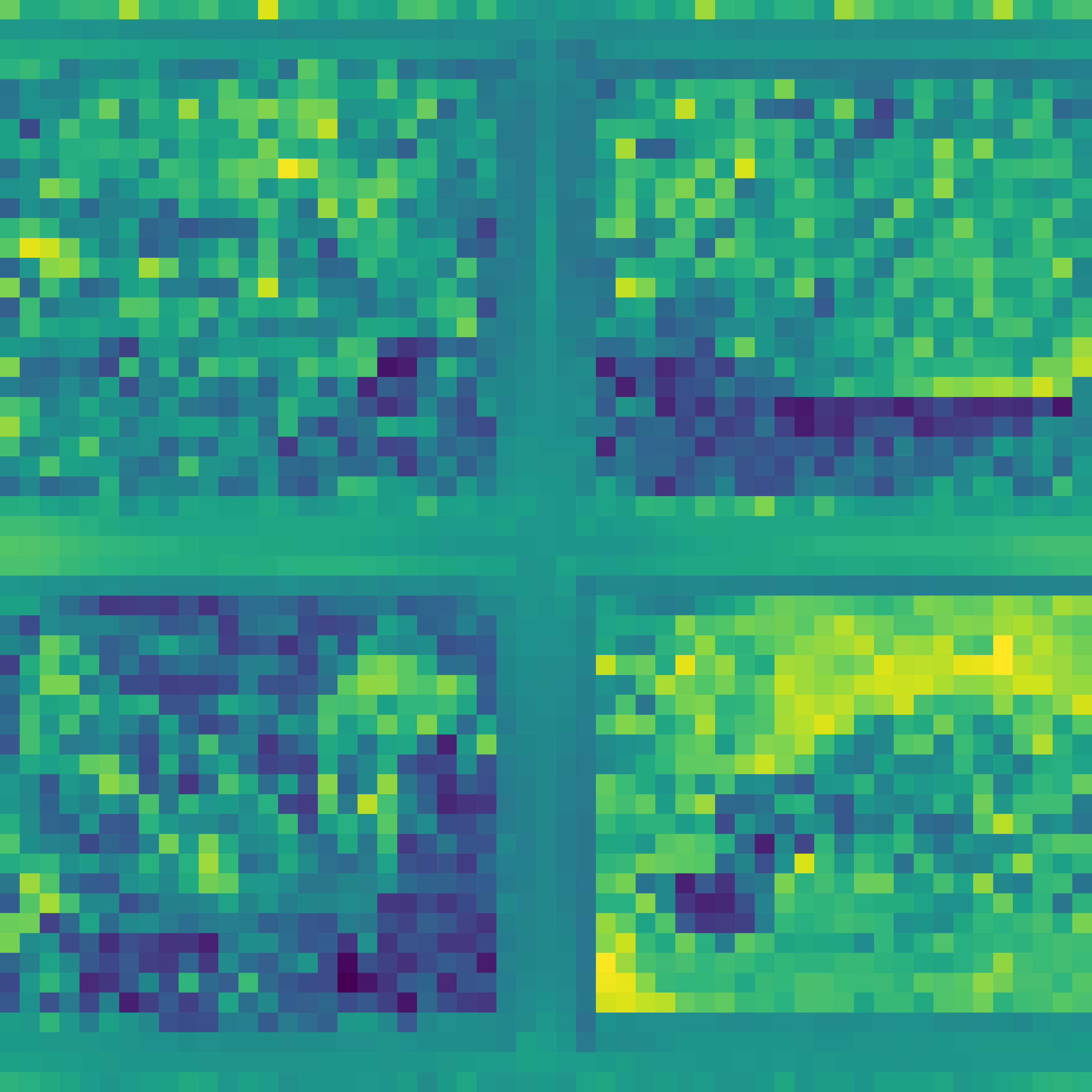} & \fig[0.09]{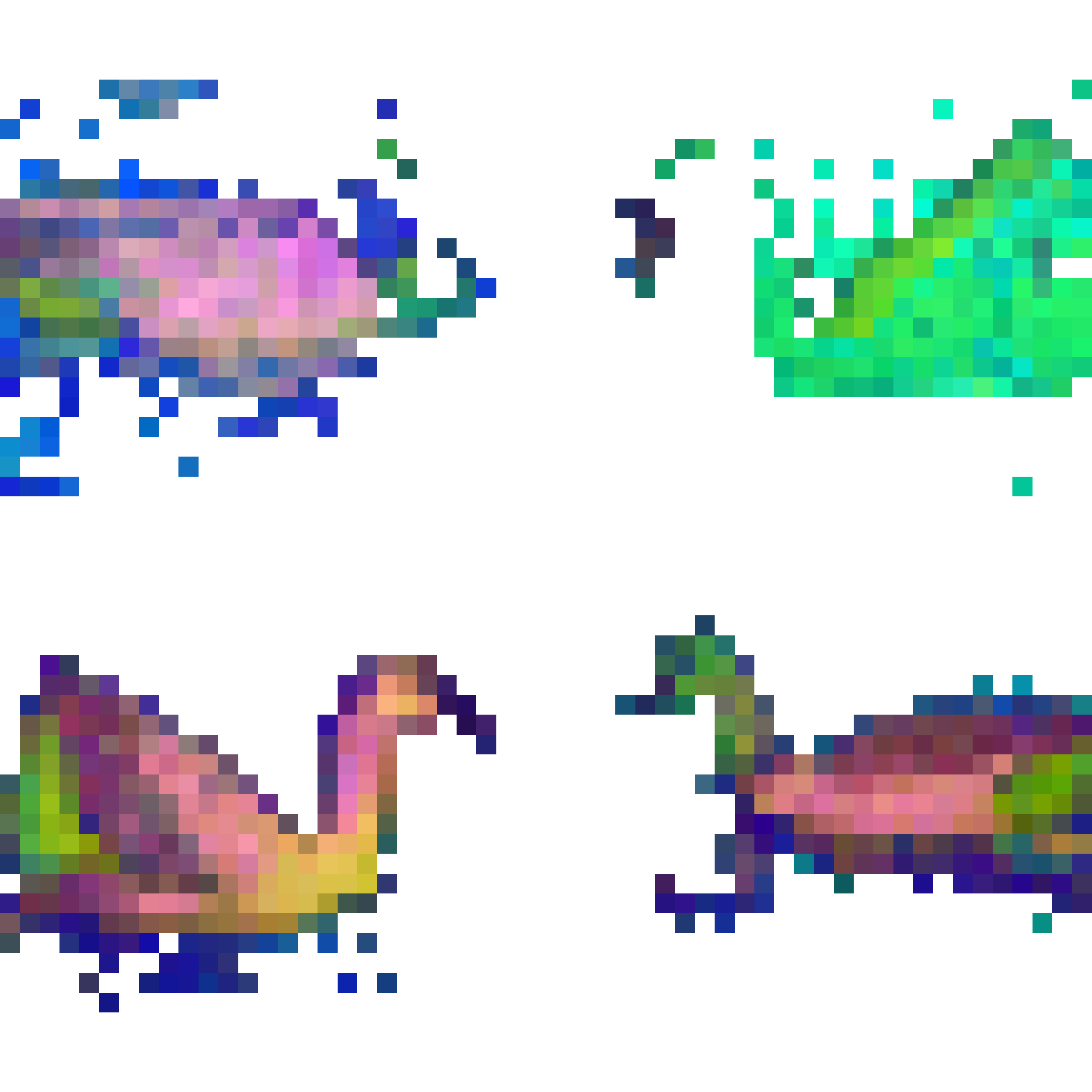} & \fig[0.09]{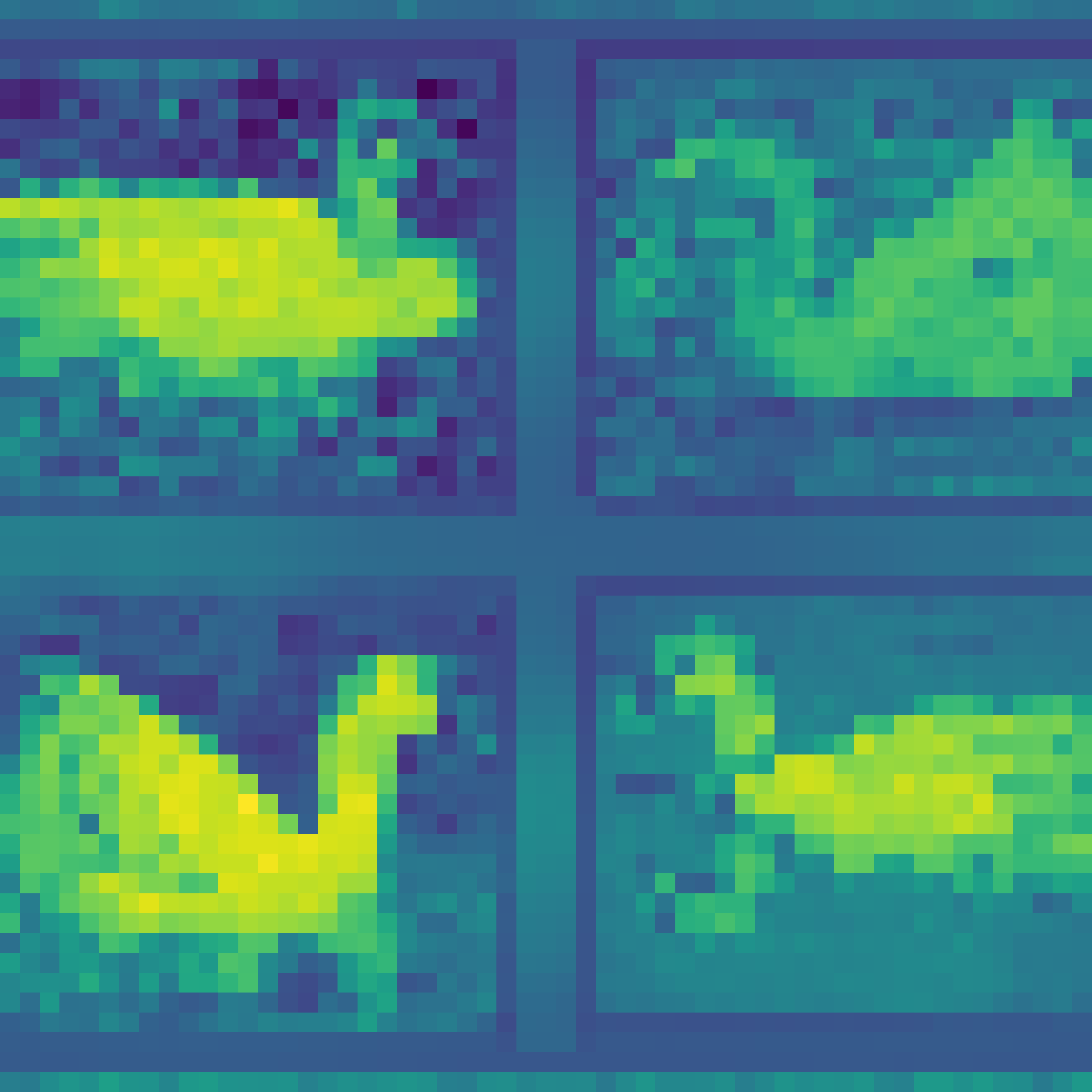} & \fig[0.09]{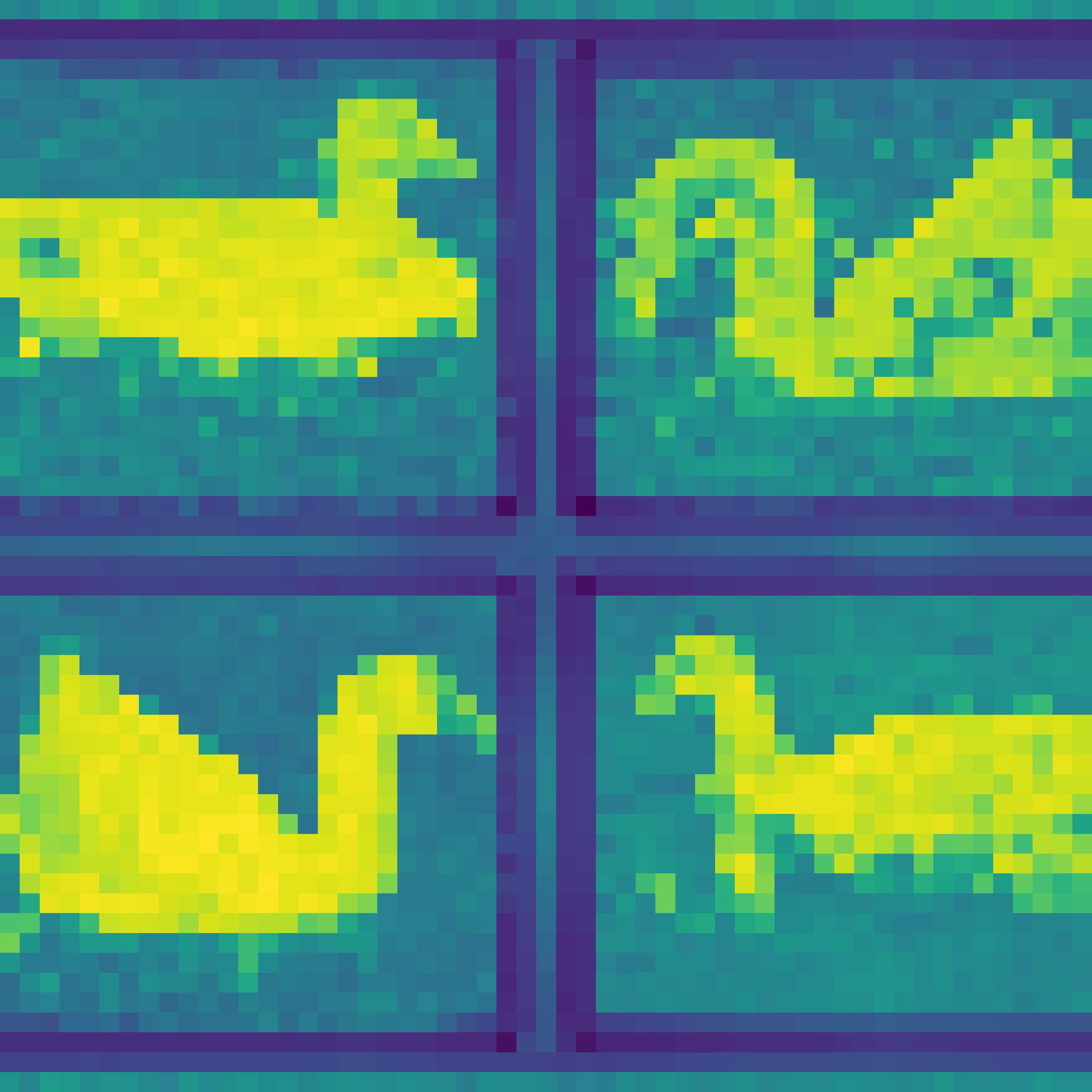} \\
    \fig[0.09]{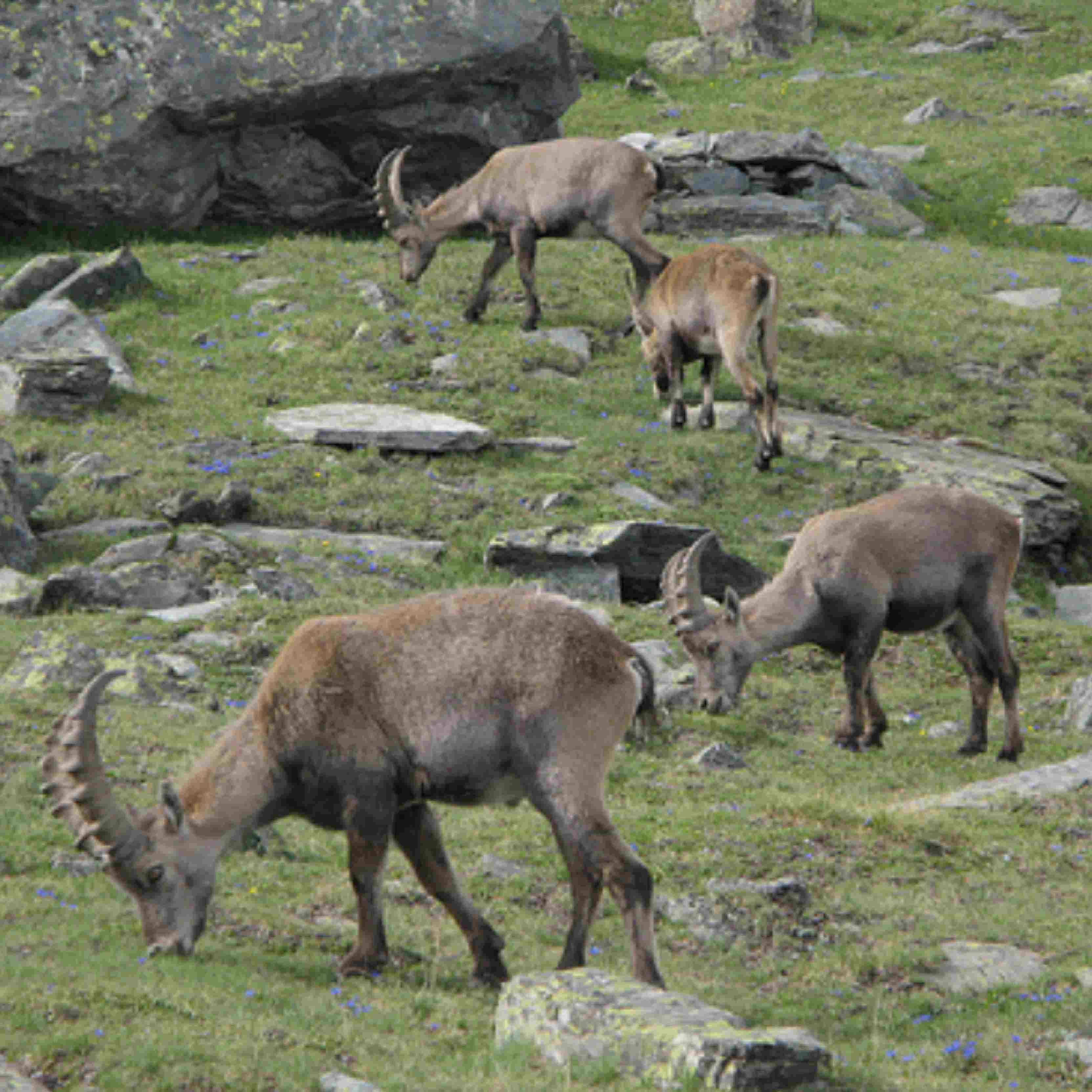} & \fig[0.09]{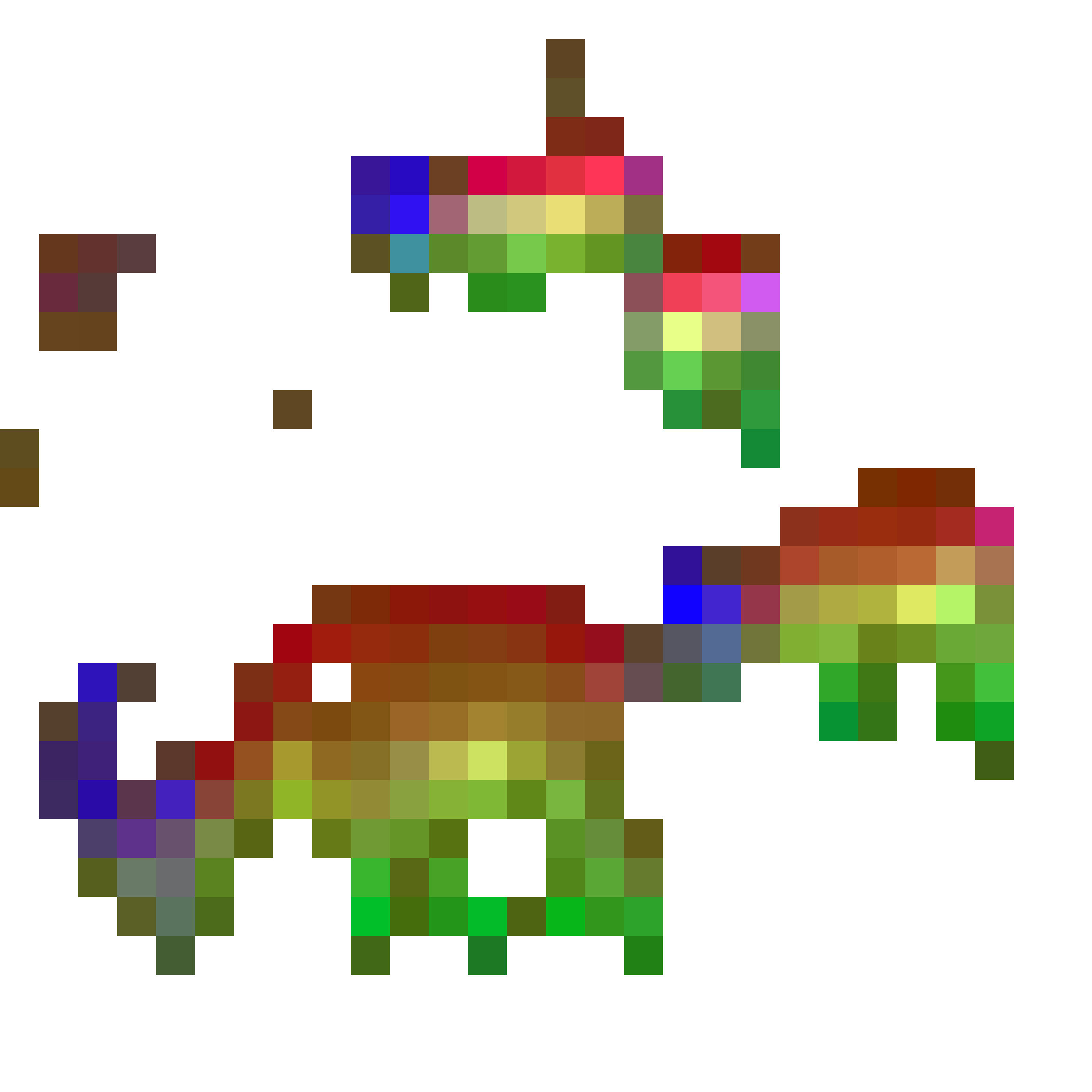}                  & \fig[0.09]{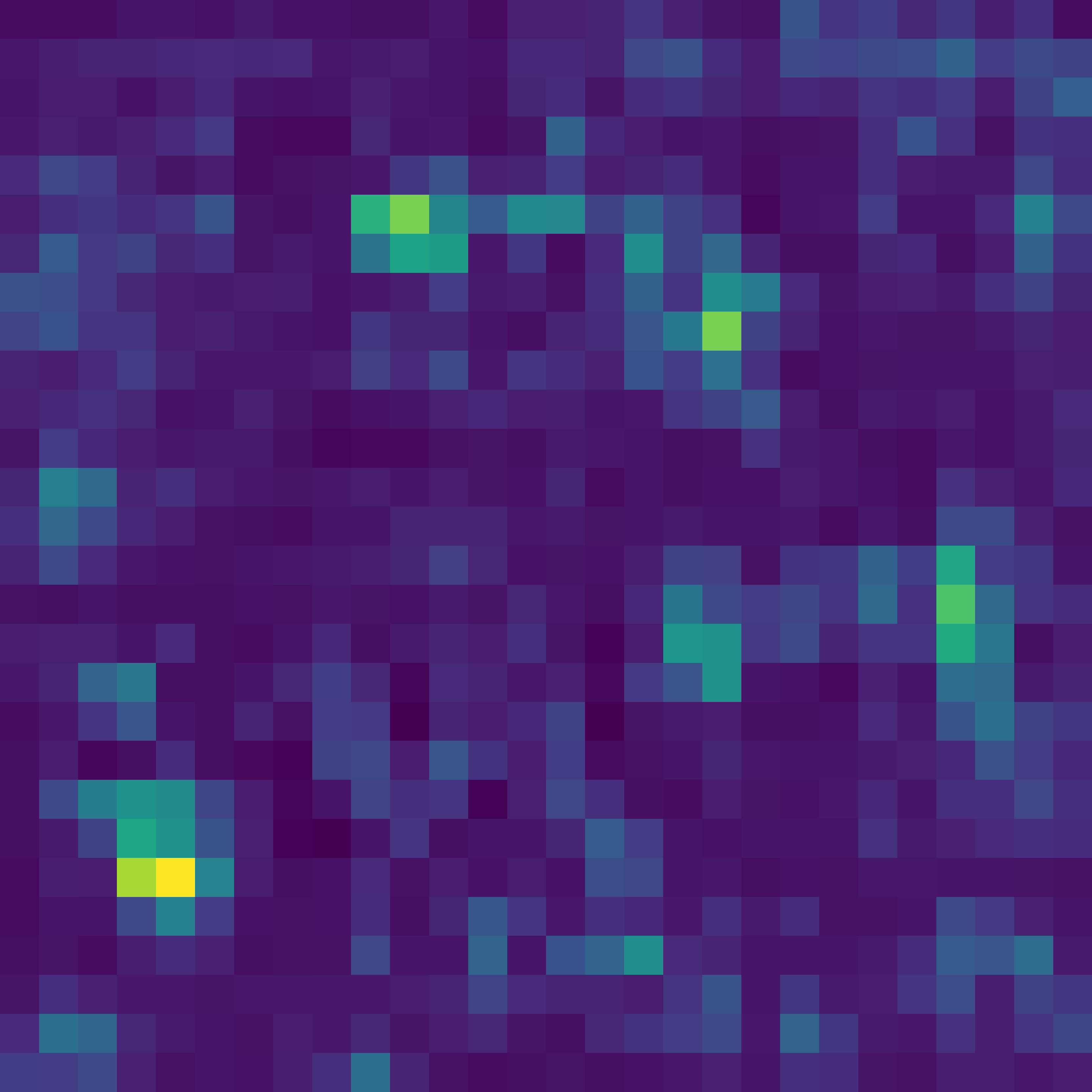}                                       & \fig[0.09]{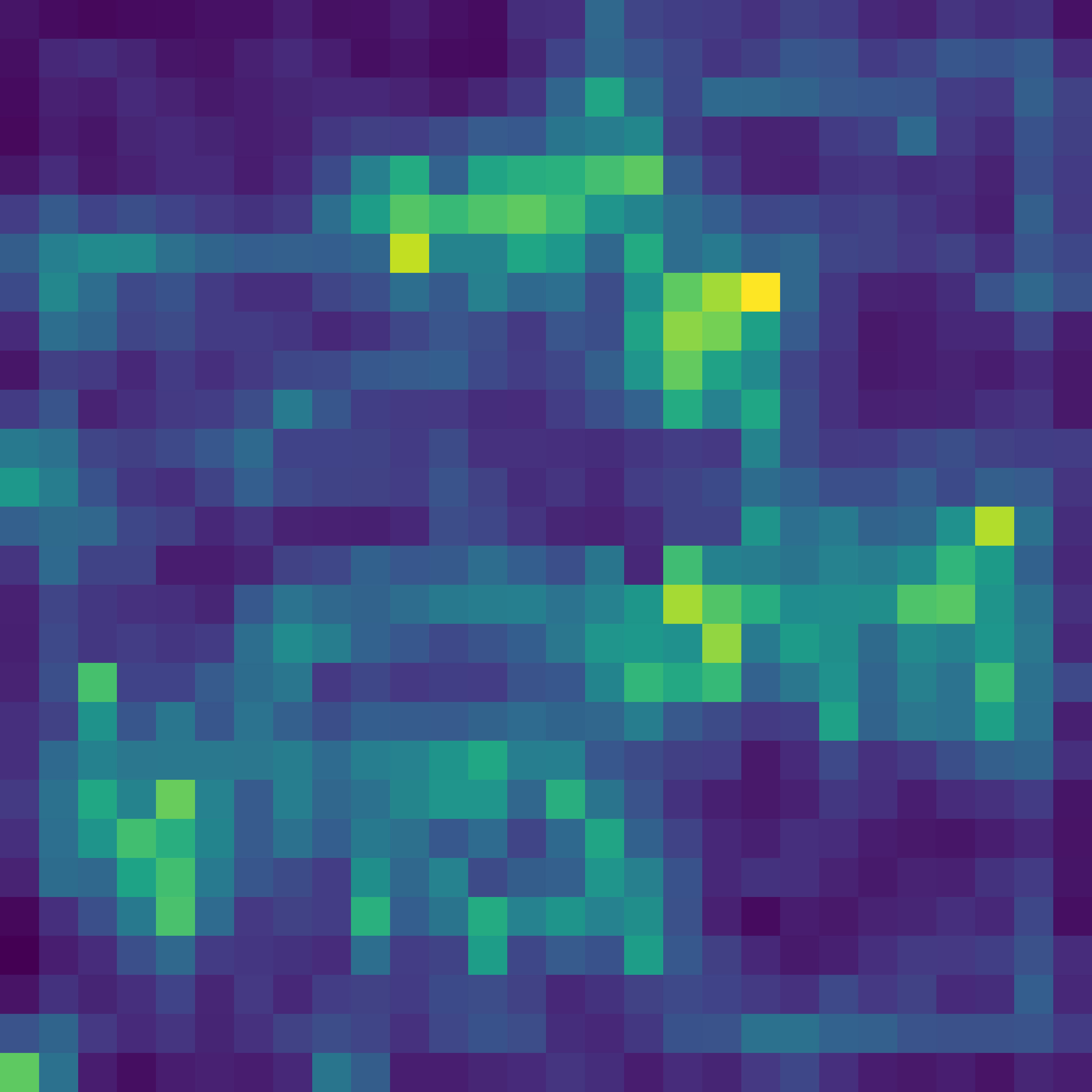}                          & \fig[0.09]{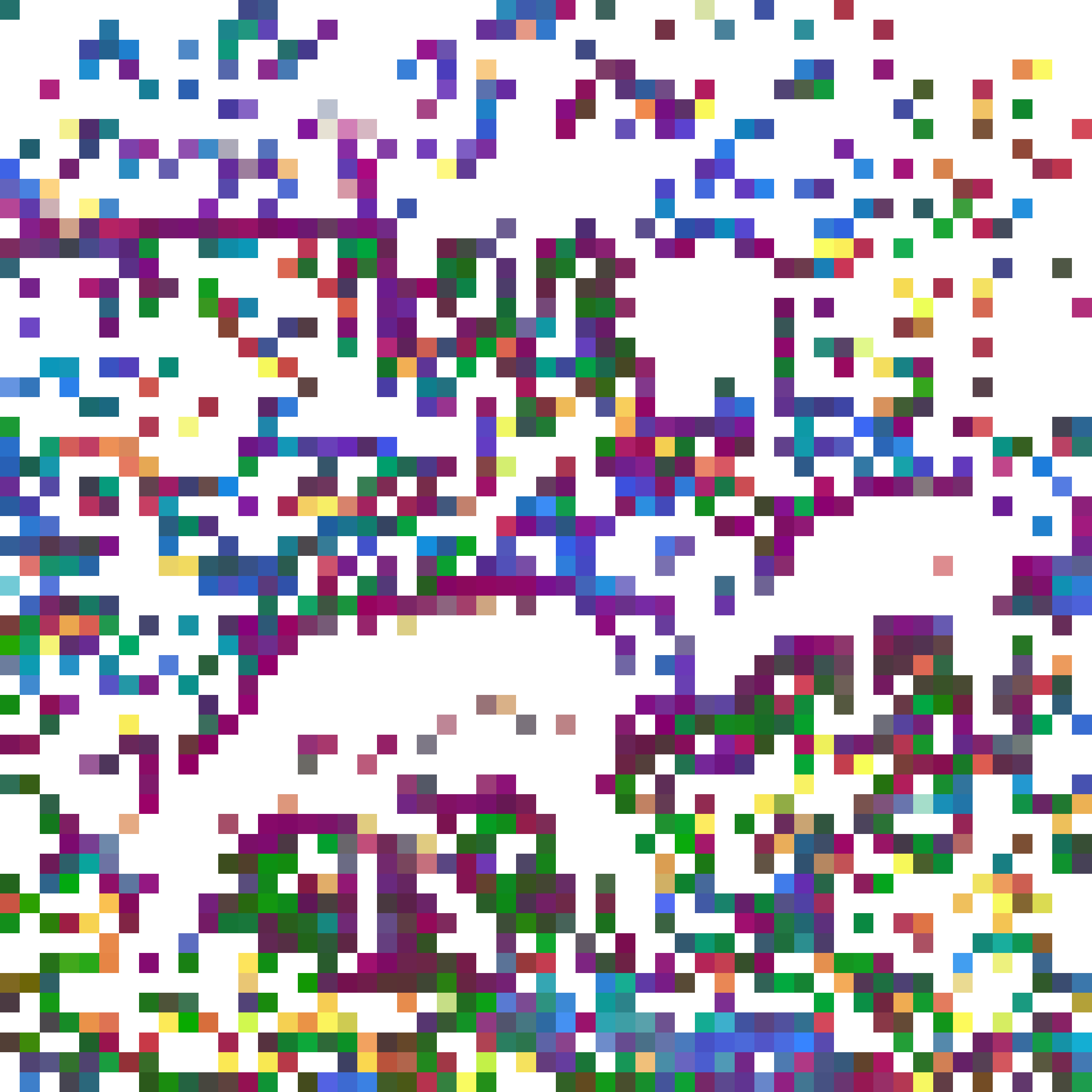} & \fig[0.09]{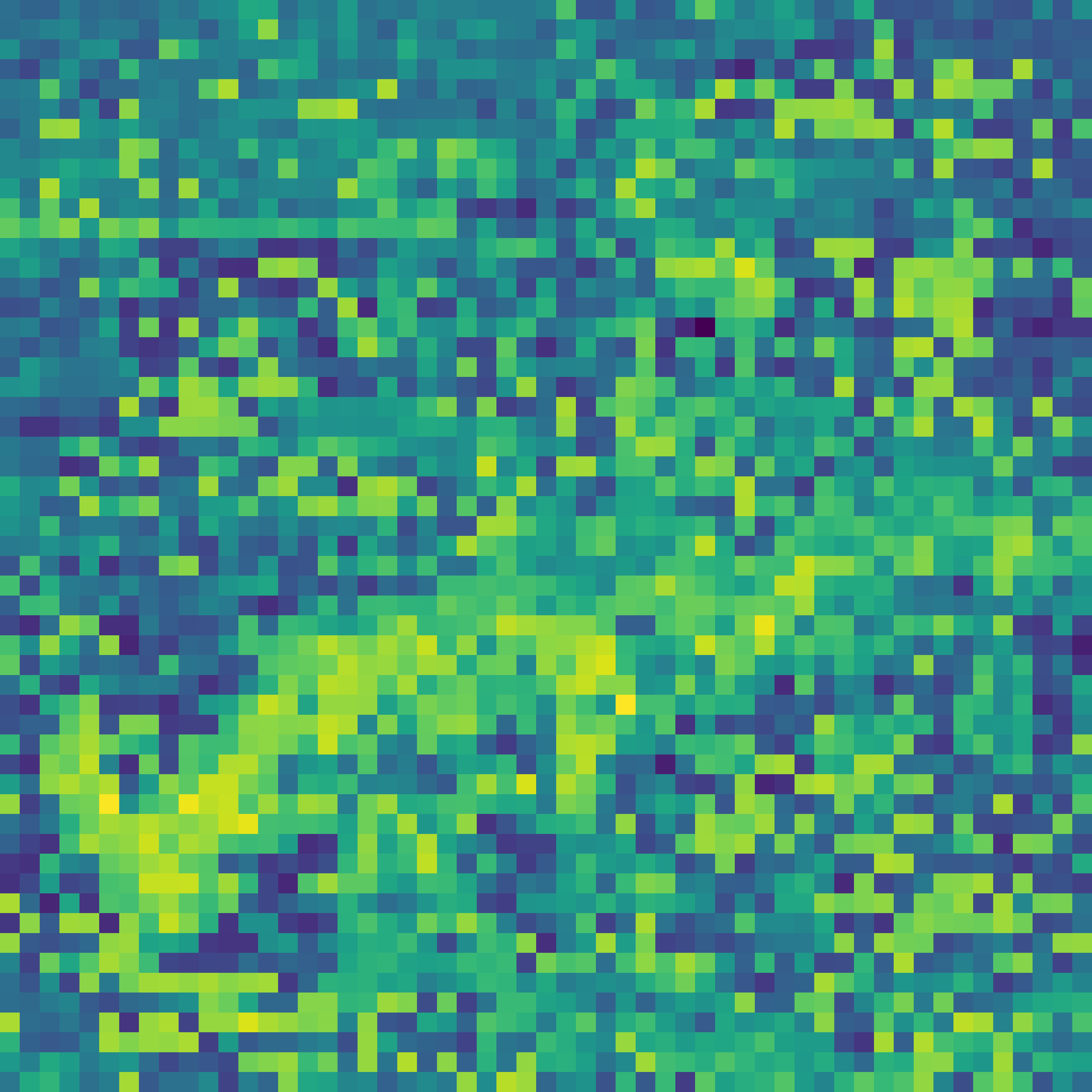} & \fig[0.09]{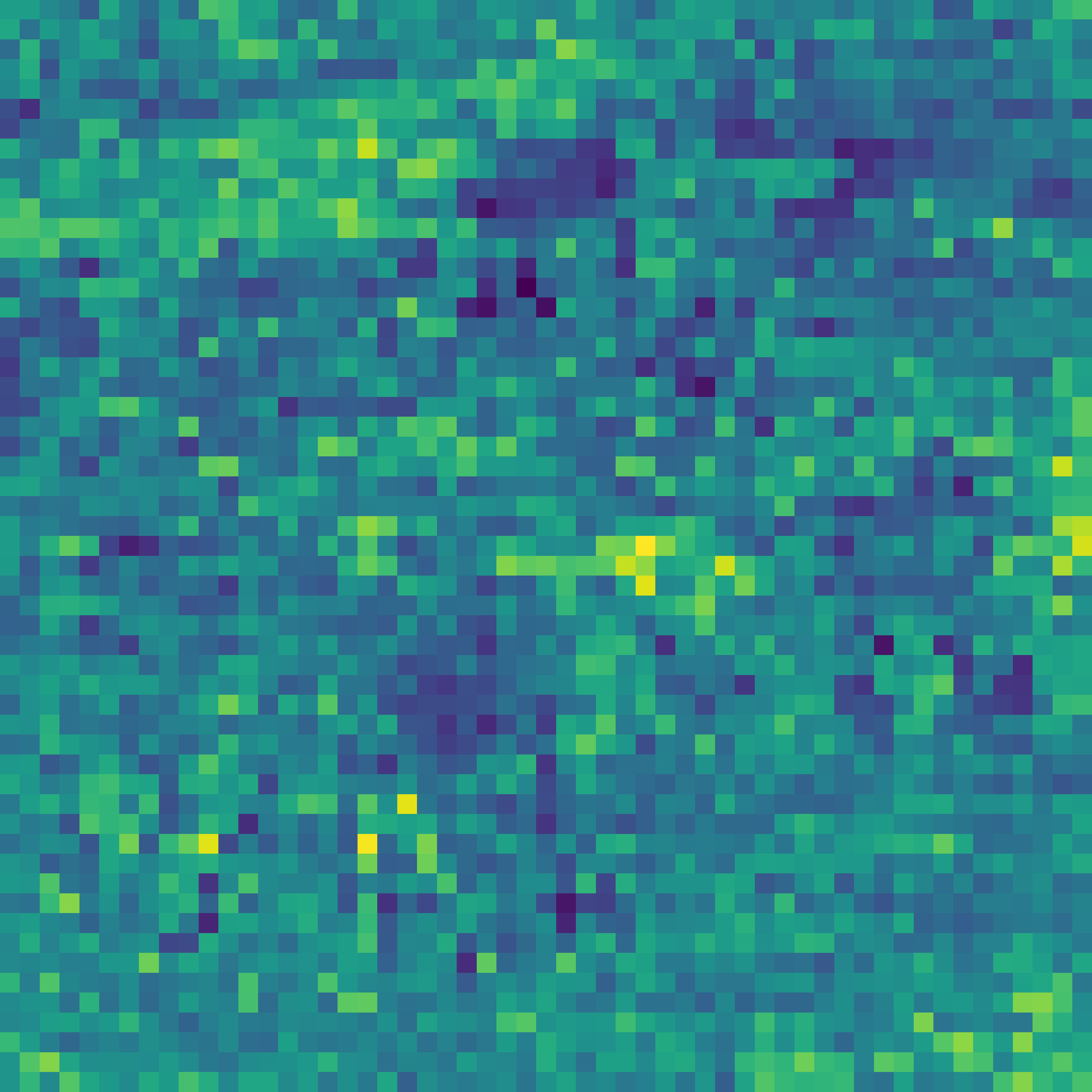} & \fig[0.09]{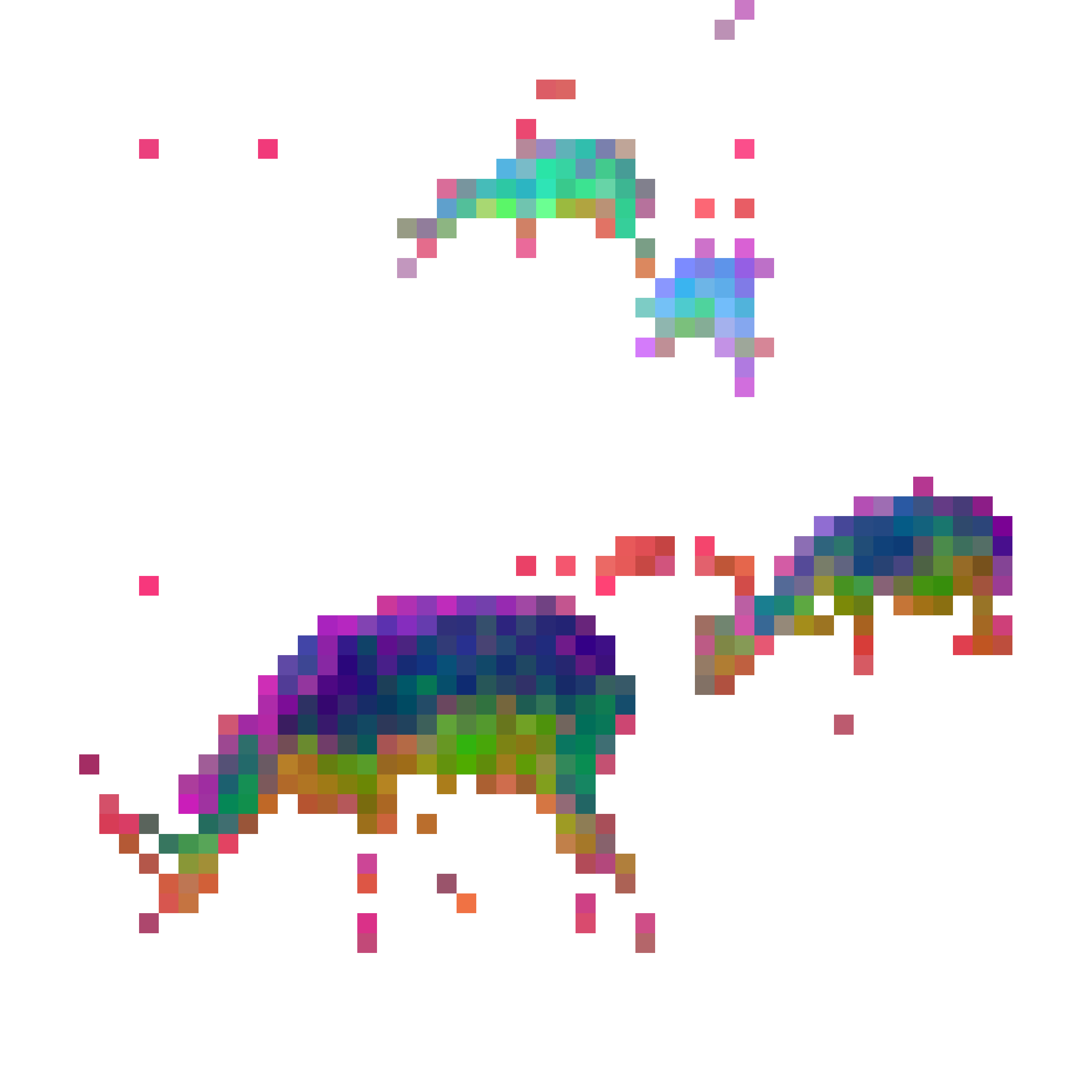} & \fig[0.09]{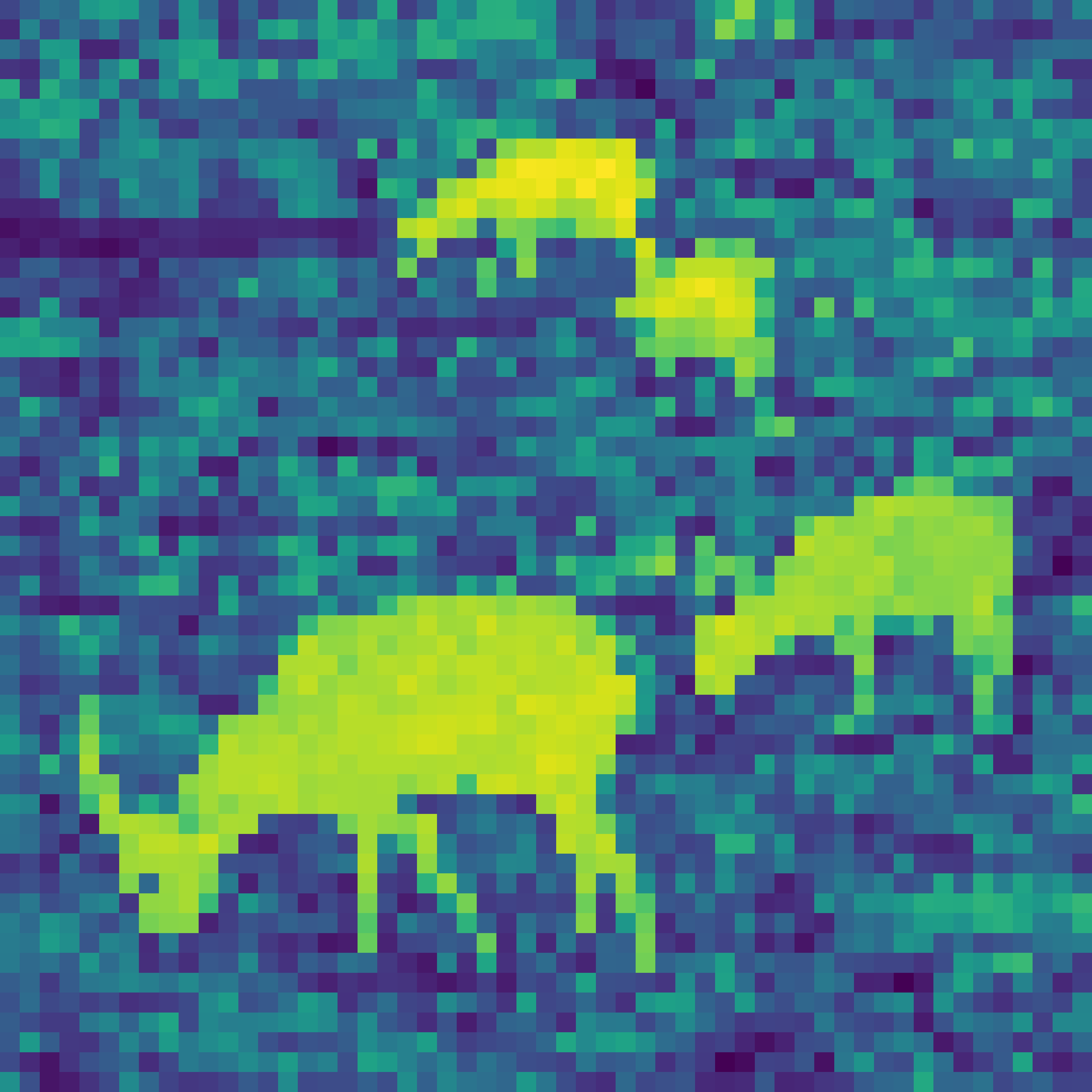} & \fig[0.09]{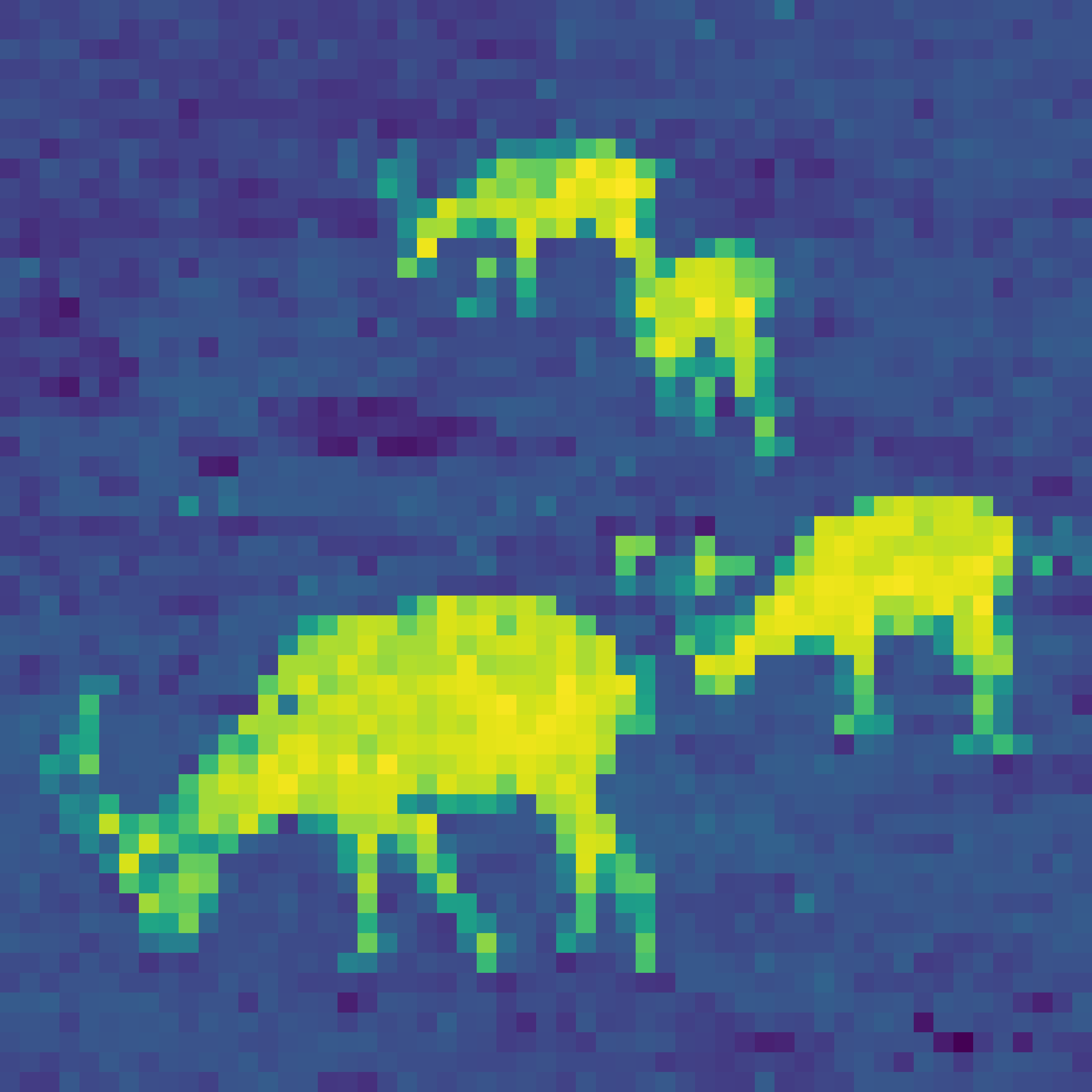} \\
\end{tabular}

\vspace{-6pt}
\caption{LazyStrike provides a unified framework for analyzing and mitigating diverse artifacts across different supervision settings in ViTs. The figure visualizes patch scores—defined as CLS–patch similarity—under full supervision (middle) and self-supervision (right), together with PCA projections of features under full supervision (left).
}
\label{fig:problem_vit}

\end{center}
}]
\renewcommand{\thefootnote}{\fnsymbol{footnote}}
\footnotetext{$^{\dagger}$ Corresponding author}
\begin{abstract}
    Vision Transformers (ViTs), when pre-trained on large-scale data, provide general-purpose representations for diverse downstream tasks. However, artifacts in ViTs are widely observed across different supervision paradigms and downstream tasks. 
    Through systematic analysis of artifacts in ViTs, we find that their fundamental mechanisms have yet to be sufficiently elucidated.
    In this paper, through systematic analysis, we conclude that these artifacts originate from a lazy aggregation behavior: ViT uses semantically irrelevant background patches as shortcuts to represent global semantics, driven by global attention and Coarse-grained semantic supervision. Our solution selectively integrates patch features into the \texttt{CLS} token, reducing the influence of background-dominated shortcuts and consistently improving performance across 12 benchmarks under label-, text-, and self-supervision.
    We hope this work offers a new perspective on ViT behavior.
    \vspace{-5mm}
\end{abstract}

\section{Introduction}
\label{sec:intro}

Vision Transformers (ViTs)~\cite{vit} have become the \textit{de facto} standard
for image recognition~\cite{deng2009imagenet,lou2025overlock, lou2025sparx,lou2025transxnet,yan2015hd,he2022attribute}. 
More importantly, they serve as general-purpose feature extractors across various
specific vision tasks~\cite{he2017mask,cheng2021mask2former,fu2025segman, zhu2025rethinking, tang2023contrastive}, functioning as a
frozen foundation model pre-trained on large-scale data to embed images into
feature representations, enabled by their scalability in data and model size.
More broadly, this generic ViT feature extractor can adapt to various supervision
methods during pre-training, with different approaches exhibiting
characteristics particularly suited to diverse downstream tasks. Specifically,
\textit{\textbf{supervised methods}}—such as training ViTs with fully-supervised
classification labels or text-supervised image–text pairs (\eg, in models like
CLIP~\cite{clip})—produce dense features for open-vocabulary tasks, and function
as visual encoders for large vision-language models (LVLMs)~\cite{liu2023llava}.
Alternatively, \textit{\textbf{self-supervised methods}}~\cite{moco,dinov1,dinov2},
particularly the DINO~\cite{dinov1} model trained solely on images, demonstrate
the potential for object and part discovery, making them applicable to unsupervised
segmentation tasks~\cite{lost}.

However, recent studies uncover puzzling \emph{dense-feature artifacts} in ViTs when applied to
downstream tasks requiring dense features. For instance, 
DINO~\cite{dino} demonstrates that label-supervised ViTs suffer from an \textbf{attention} deficit~\cite{simpool}, while CLIPSelf~\cite{wu2023clipself} observes that text-supervised ViTs fail to produce dense image \textbf{features} that are accurately aligned with textual cues in open-vocabulary tasks. Meanwhile, Register~\cite{registers} reveals that self-supervised ViTs generate artifacts in the \textbf{attention} maps, commonly referred to as high-norm tokens, which adversely affect object localization tasks~\cite{lost}.

These phenomena suggest a common underlying issue in ViTs, merely manifesting differently under various supervision paradigms~\cite{dinov1,deit,openclip}. 
In our preliminary exploration, we found that no single method~\cite{registers,sclip,maskclip} could comprehensively address these phenomena.
This result suggests that our understanding of ViTs remains incomplete, even though they have been studied for roughly half a decade~\cite{vit}.
Given that many issues may stem from a shared mechanism, a unified solution is desirable. 
In this paper, we undertake a first-principles investigation – systematically defining, analyzing, and addressing the different types of artifacts observed in ViTs from the ground up.

To establish a unified definition for these phenomena across different paradigms, we introduce the \textbf{Patch Score}—the similarity between patch features and the \texttt{CLS} token, which encapsulates an image’s global semantics—thereby assessing local semantic consistency relative to the global representation, independent of the training paradigm.
The intuition behind Patch Score is that for ViT under different supervision, the training objective aims to align the \texttt{CLS} feature with supervisory signals (\eg, labels or text); any misalignment in dense features results in increased Patch Scores in non-foreground regions, as shown in Fig.~\ref{fig:problem_vit}. 
To quantitatively assess artifacts in Patch Scores, we propose the \textbf{Point-in-Box (PiB)} metric, which evaluates whether the patch with the highest score lies within the annotated foreground region.
As shown in Fig.~\ref{fig:problem_vit} and Tab.~\ref{tab:pib}, we find that across different supervision settings, ViTs assign higher Patch Scores to background patches and achieve much lower PiB compared with ConvNets~\cite{resnet}. Detailed experimental settings are provided in Sec.~\ref{sub:ps}.
For clarity and simplicity, unless explicitly stated otherwise, the term ``artifacts'' in the following text specifically refers to semantically irrelevant background tokens that erroneously yield high Patch Scores.

Based on Patch Score and PiB, we conduct an in-depth analysis into ViT's behavior and propose a
hypothesis aimed at better explaining these artifacts:

\begin{table}[t]
    \centering
    \begin{adjustbox}
        {max width=0.93\linewidth}
        \begin{tabular}{lcc}
            \toprule \textbf{Method}                                                 & \textbf{High Norm}                   & \textbf{Point-in-Box (PiB)}      \\
            \midrule \textcolor{gray}{ResNet~\cite{resnet}}                 & \textcolor{gray}{\ding{55}} & \textcolor{gray}{68.4} \\
            ViT~\cite{vit}                                                  & \ding{51}                   & 42.7                   \\
            \hspace{0.7em}+Register~\cite{registers}                        & \ding{55}                   & 41.5                   \\
            \midrule \textcolor{gray}{DINO-ResNet~\cite{dinov1}}         & \textcolor{gray}{\ding{55}} & \textcolor{gray}{71.1} \\
            DINO-ViT~\cite{dinov1}                                           & \ding{55}                   & 45.3                   \\
            \midrule \textcolor{gray}{OpenCLIP-ResNet~\cite{clip}} & \textcolor{gray}{\ding{55}} & \textcolor{gray}{53.9} \\
            OpenCLIP-ViT~\cite{clip}                                                & \ding{51}                   & 39.8                   \\
            \hspace{0.7em}+Register~\cite{registers}                        & \ding{55}                   & 37.6 \\
            \bottomrule
        \end{tabular}
    \end{adjustbox}
    \caption{Point-in-Box (PiB) across different supervision methods. We find that
    Register reduces high-norm tokens but high-norm is not the root cause of artifacts.  }
    \vspace{-8pt}
    \label{tab:pib}
\end{table}

\begin{itemize}
    \setlength{\itemsep}{0pt}
    \setlength{\parsep}{0pt}
    \setlength{\parskip}{0pt}


    \item Natural images inherently contain many background patches that are irrelevant to the primary object. 
With only image-level supervision, the model lacks spatial guidance and thus tends to encode global semantics via background evidence (lazy aggregation). 
Empirically, removing the top 50\% highest-scoring patches in a pre-trained ViT has negligible impact on ImageNet accuracy (Fig.~\ref{fig:problem}), corroborating this reliance.

    \item
        \textit{Global dependencies allow ViT to exploit these extraneous background patches as shortcuts to represent global semantics.}
        In the absence of patch-level annotations, ViTs may adopt a lazy aggregation by diffusing small foreground semantics to background at the beginning of training (Fig.~\ref{fig:stage}).
        We validate that reducing global dependencies indeed mitigates artifact phenomena (Tab.~\ref{app:tab:hyper}).
\end{itemize}

Building on this interpretation, we further validate our hypothesis by proposing
a straightforward solution to eliminate these artifacts:
\textit{By regulating the influence of background patches during pre-training, we encourage ViTs to focus on foreground semantics.} 
Specifically, the model learns to estimate the contribution of each token (Sec.~\ref{sub:fouri}) and selectively integrate informative patch features into the \texttt{CLS} token to strengthen foreground representation.
As shown in Fig.~\ref{fig:cls_after_lazystrike}, ViTs automatically shift their attention to
foreground objects, aligning high-scoring patches with the foreground as these ratios
are appropriately increased. Once this lazy aggregation is mitigated, our approach – termed LaSt-ViT (LazyStrike ViT)– eliminates Patch Score artifacts across all types of supervision. It effectively addresses both the high-norm token issue and feature misalignment.
Notably, after applying our method, ViTs exhibit improved emergent semantic segmentation properties~\cite{dino} consistently across different pre-training paradigms.

\textbf{Contributions.}
(1) We systematically analyze the root cause of artifacts in ViTs via \emph{Patch Score} and \emph{Point-in-Box (PiB)}, revealing a background-dominant bias that emerges early and persists. 
(2) We provide a hypothesis linking \emph{Coarse-grained semantic supervision} and \emph{global dependencies} to \emph{lazy aggregation}—a shortcut behavior where ViTs rely on background patches to encode global semantics instead of attending to true foreground regions.
(3) We propose LaSt-ViT, a simple, frequency-aware selective aggregation scheme that anchors the CLS token to foreground regions. 
(4) We demonstrate consistent gains across 12 benchmarks including object discovery, semantic/instance segmentation, and open-vocabulary detection.

\section{Related Work}
\label{sec:rw}


\noindent \textbf{Artifacts in text-supervised ViTs (CLIP-type models~\cite{clip}).} 
Recent advances in vision–language contrastive pretraining~\cite{clip,align} have enabled CLIP models to produce dense predictions beyond image-level classification. 
MaskCLIP~\cite{maskclip} first showed that CLIP features can yield zero-shot semantic segmentation via pixel–text alignment. 
However, later studies~\cite{sclip,proxyclip,corrclip,naclip,clipsurgery,wu2023clipself} found that although ViTs surpass ResNets in model capacity and classification accuracy, they perform worse on dense alignment tasks. 
To mitigate this misalignment, existing works either (1) modify the final attention layers~\cite{sclip,proxyclip,corrclip,naclip,clipsurgery,clipdinoiser} or (2) introduce additional alignment training~\cite{wu2023clipself,chen2025vision} or (3) employ test-time activation editing, such as dynamically shifting high-norm outlier activations into untrained register tokens during inference~\cite{jiang2025vision}. In contrast, our method tackles the problem directly during pretraining, avoiding architectural changes, post-hoc fine-tuning, and test-time interventions, fundamentally preventing the emergence of lazy behavior in text-supervised ViTs.

\noindent \textbf{Artifacts in self-supervised ViT (DINO-type model~\cite{dinov1})}. 
Register~\cite{registers} found that DINOv2~\cite{dinov2} leads to successful monocular depth estimation and semantic segmentation, but it loses the object detection capability of DINO~\cite{dinov1} due to artifacts appearing on the feature map. To address this issue, additional tokens were introduced, designed to store global features and mitigate the impact of these artifacts.
During our in-depth analysis of the high-norm phenomenon, we found that high-norm is merely a manifestation of the lazy behavior in later stages. Simply moving the high-norm tokens from the feature map to the register tokens does not fully address the underlying deficiencies in downstream tasks. Therefore, \textit{Vision Transformer requires more than just Registers.}


\section{Preliminary}

\subsection{Network Architecture: Vision Transformer}
Given an image $\mathbf{x}\in\mathbb{R}^{H\times W\times 3}$, ViTs~\cite{vit} split it into non-overlapping $P\times P$ patches and linearly project them to
$\mathbf{x}_{\text{emb}}\in\mathbb{R}^{N\times D}$ with $N=\tfrac{HW}{P^2}$ via the patch embedding $\mathcal{P}_{\text{emb}}(\cdot)$. 
An encoder $\mathcal{P}_{\text{enc}}(\cdot)$ consisting of a stack of transformer blocks updates tokens with self-attention. 
Global aggregation is applied to the output of the encoder using one of two standard forms:
\begin{align}
     \mathbf{x}_{\text{patch}} = \mathcal{P}_{\text{enc}}( \mathcal{P}_{\text{emb}}(\mathbf{x})), & \mathcal{Q}_{\text{CLS}} = \textit{Pooling}(\mathbf{x}_{\text{patch}}),\label{equ1} \tag{1} \\
      \textit{o}&\textit{r} \notag{}\\
      \mathbf{x}_{\text{patch}}, \mathcal{Q}_{\text{CLS}} =  \mathcal{P}_{\text{enc}}&( \mathcal{P}_{\text{emb}}(\mathbf{x}), \mathcal{O}_{\text{CLS}}),\label{equ2} \tag{2} 
 \end{align}
where \textit{Pooling} is global average pooling (GAP) over patch tokens in \cref{equ1}; 
$\mathcal{O}_{\text{CLS}}$ is a learnable query concatenated before encoding in \cref{equ2}, whose output token $\mathcal{Q}_{\text{CLS}}$ serves as the global representation.

\section{Analysis and Hypothesis}
\label{sec:method}

We introduce two probes—\emph{Patch Score} (CLS–patch similarity) and \emph{Point-in-Box}—to analyze where and when artifacts emerge in ViTs (Sec.~\ref{sub:ps}). 
From both spatial and temporal perspectives (Sec.~\ref{sub:artifacts}), we find that high patch scores concentrate in background regions, and this bias appears from the very beginning of training and persists throughout. 
These findings suggest that ViTs, trained with \emph{Coarse-grained semantic supervision} (image-level rather than patch-level objectives) and equipped with strong \emph{global dependencies} (long-range attention), tend to adopt a \emph{lazy aggregation} shortcut—diffusing foreground semantics into background tokens. 
To verify this hypothesis, we isolate each factor: reducing background tokens by using a larger \emph{patch size} in the embedding layer (Sec.~\ref{sub:sparsity}) and constraining the attention range via window-based attention (Sec.~\ref{sub:window}).
Both interventions raise the Point-in-Box score but slightly lower classification accuracy, indicating that reducing global dependencies helps suppress background bias at the cost of overall recognition performance.
All analyses are conducted on ImageNet-1k~\cite{deng2009imagenet,deit}, with consistent trends observed under text- and self-supervised pretraining~\cite{openclip,dinov1,laion}. 
Further observations—such as the role of high-norm tokens~\cite{registers} and the unique behavior of DINO-v1—are provided in the Appendix.

\subsection{New Metric: Patch Score and Point-in-Box}
\label{sub:ps}
\noindent \textbf{Patch Score.}
To enable a unified comparison across architectures and pretraining settings, we define the \emph{Patch Score} as the similarity between each patch and the global representation. 
For ViTs, the global representation is the \texttt{CLS} token $\mathcal{Q}_{\text{CLS}}$; 
for ConvNets, it is the feature after global average pooling $\mathcal{Q}_{\text{GAP}}$, which serves as an implicit \texttt{CLS} token.
Formally,
\begin{align}
\mathcal{S}_{\text{p}} = 
\frac{\mathbf{x}_{\text{patch}} \cdot Q_{\text{CLS}}}
{\|\mathbf{x}_{\text{patch}}\|_2 \, \|Q_{\text{CLS}}\|_2}, \tag{3}
\end{align}
where higher Patch Scores indicate stronger alignment with image-level semantics.

\noindent \textbf{Point-in-Box benchmark.} Building on the patch score, we assess artifacts by determining whether the highest scoring regions correspond to foreground objects. We use images from the ImageNet~\cite{deng2009imagenet} validation set that feature a single object annotation to avoid ambiguity. We define the Point-in-Box score as the proportion of images where the highest patch score falls within the foreground bounding box.

\subsection{Artifacts in Patch Score}
\label{sub:artifacts}
\useqaA
\begin{question}
Where does the \texttt{CLS} token ``look'' at?
\end{question}

\noindent \textbf{\textcolor{black}{Experiment Setting.}} 
We study a ViT-B/16 trained on ImageNet-1k~\cite{deng2009imagenet} (fully supervised). 
We visualize the normalized patch-score distributions, and perform a probe by masking the top-$k$ or bottom-$k$ patches directly on the input image prior to re-evaluation.

\begin{figure}[h]
\scriptsize
\centering
\includegraphics[width=0.5\textwidth]{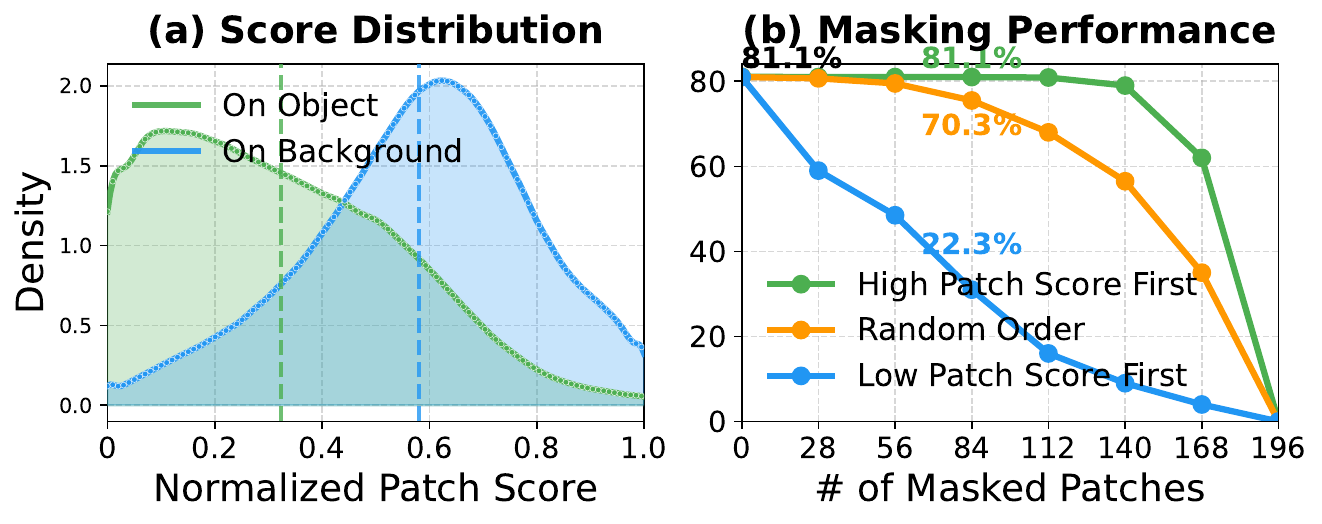}
\vspace{-8pt}
\caption{\textbf{Patch‑score distribution and masking probe on ImageNet‑1k.} (a) Normalized distributions of patch scores for foreground vs. background. (b) Removing top-$k$ high‑score patches (up to $70\%$) does not hurt accuracy and even can improve it. 
}
\vspace{-4pt}
\label{fig:problem}
\end{figure}

\noindent \textbf{\textcolor{black}{Experiment Results.} } The experimental results show that:
\begin{itemize}[leftmargin=10pt]
\setlength{\itemsep}{0pt}
\setlength{\parsep}{0pt}
\setlength{\parskip}{0pt}
    \item \textbf{Distribution.} 
    Foreground patches concentrate at lower patch-score values, while background patches dominate the high-score tail (Fig.~\ref{fig:problem}\textcolor{cvprblue}{a}).
    \item \textbf{Masking Probe.} 
    Removing high-score patches does not harm accuracy—and can even slightly improve it (e.g., +1.2\% for ViT-B/16)—even when more than 50\% of patches are masked. 
    In contrast, removing low-score patches leads to a sharp accuracy drop (up to 60\% at 70\% masking; Fig.~\ref{fig:problem}\textcolor{cvprblue}{b}).
\end{itemize}

\begin{answer}
Masking out background patches in the input image barely affects classification performance, 
yet these regions show feature responses strongly correlated with the \texttt{CLS} token that encodes class semantics, 
suggesting that foreground information has been propagated into the background during training—
a behavior likely driven by the dominance of background tokens over foreground ones in natural images (as shown in Fig.~\ref{fig:problem}\textcolor{cvprblue}{b}, where over half of the patches are non-contributive to classification).
\end{answer}

\useqaB
\begin{question}
When does this phenomenon begin?
\end{question}

\noindent \textbf{\textcolor{black}{Experiment Setting.}}
We train ViT-B/16~\cite{vit} and ResNet-50~\cite{resnet} on ImageNet-1k~\cite{deng2009imagenet} with identical hyperparameters and batch size, and track both \emph{top-1 accuracy} and the \emph{Point-in-Box} score throughout training.

\begin{figure}[h]
\scriptsize
\centering
\includegraphics[width=0.48\textwidth]{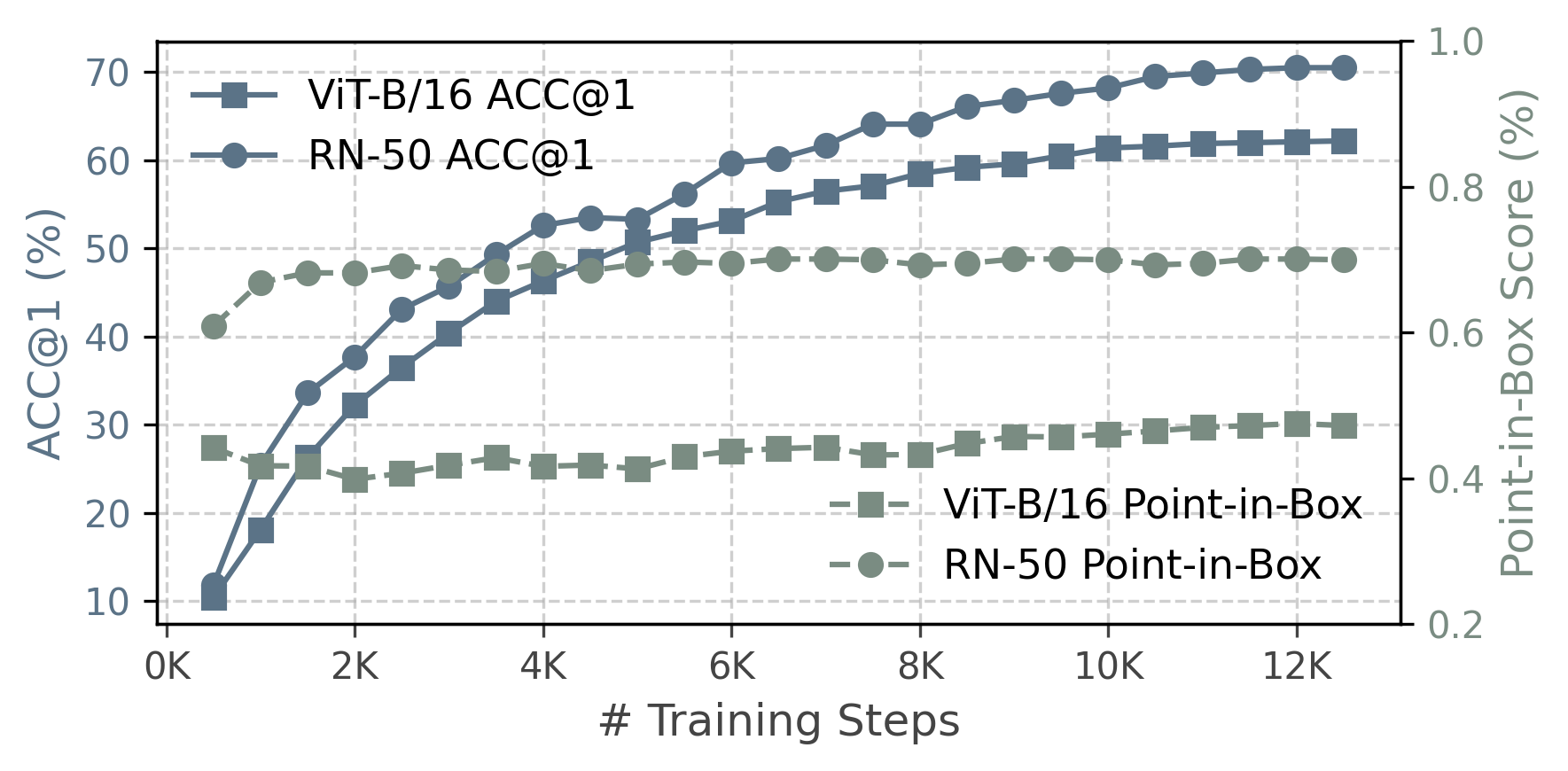}
\vspace{-8pt}
\caption{\textbf{Training dynamics on ImageNet-1k.}
Left: top-1 accuracy; Right: Point-in-Box score.
As training proceeds, ViT’s classification accuracy steadily improves, yet its Point-in-Box score remains nearly flat (around $0.42 \rightarrow 0.44$) and consistently lower than ResNet’s across the entire training process.}
\vspace{-4pt}
\label{fig:stage}
\end{figure}

\noindent \textbf{\textcolor{black}{Experiment Results.}} The experimental results show that:
\begin{itemize}[leftmargin=10pt]
\setlength{\itemsep}{0pt}
\setlength{\parsep}{0pt}
\setlength{\parskip}{0pt}
  \item \textbf{Point-in-Box dynamics.}  
The Point-in-Box score of ViT, reflecting artifact level (lower indicates stronger background bias), 
stays low and nearly unchanged during training, even as classification accuracy improves (Fig.~\ref{fig:stage}).

\item \textbf{Comparison with ResNet.}  
Compared with ResNet, ViT consistently shows a lower Point-in-Box score, revealing a more pronounced background bias despite similar image-level accuracy (Fig.~\ref{fig:stage}).

\end{itemize}

\begin{answer}
Propagation of foreground information into the background occurs from the very beginning of training.  
Even at early iterations, the \texttt{CLS} token already attends predominantly to background patches instead of foreground regions, and this bias persists throughout training—yielding accurate image-level predictions but poor patch-level alignment.
\end{answer}

This early emergence indicates that the artifacts are not late-stage byproducts but intrinsic phenomena during ViT training.  
We hypothesize that at the start of training, the \texttt{CLS} token seeks the easiest path to minimize the image-level loss, quickly learning to aggregate background tokens that correlate with the image-level label.
As a result, image-level semantics are “short-circuited” through background regions, leading to high classification accuracy but poor patch-level alignment—a hallmark of the model’s \textit{lazy aggregation} behavior.
We next hypothesize that this behavior originates from two interacting factors: (1) \textbf{Coarse-grained semantic supervision}, where image-level labels cannot provide accurate patch-level supervision; and (2) \textbf{Global dependencies}, where attention-based token mixing allows background tokens to absorb foreground information.  
Sections~\ref{sub:sparsity} and~\ref{sub:window} further isolate and quantify the contribution of each factor.

\subsection{Coarse-grained Semantic Supervision}
\label{sub:sparsity}

\noindent \textbf{\textcolor{black}{Validation Experiment Setting.}}
To evaluate the effect of coarse-grained semantic supervision, we reduce the prevalence of background tokens by increasing the \emph{patch size} used in the embedding module $\mathcal{P}_{\text{emb}}(\cdot)$. 
As the patch size grows, fewer tokens are generated, and many small background regions are merged into larger patches, thereby reducing the relative proportion of background tokens.
Specifically, we train ViT-Base on ImageNet-1k~\cite{deng2009imagenet} with a $28{\times}28$ patch size (default: $16{\times}16$), which decreases the proportion of background tokens by about $10\%$ (see Appendix for details).

\begin{figure}[h]
\scriptsize
\centering
\includegraphics[width=0.48\textwidth]{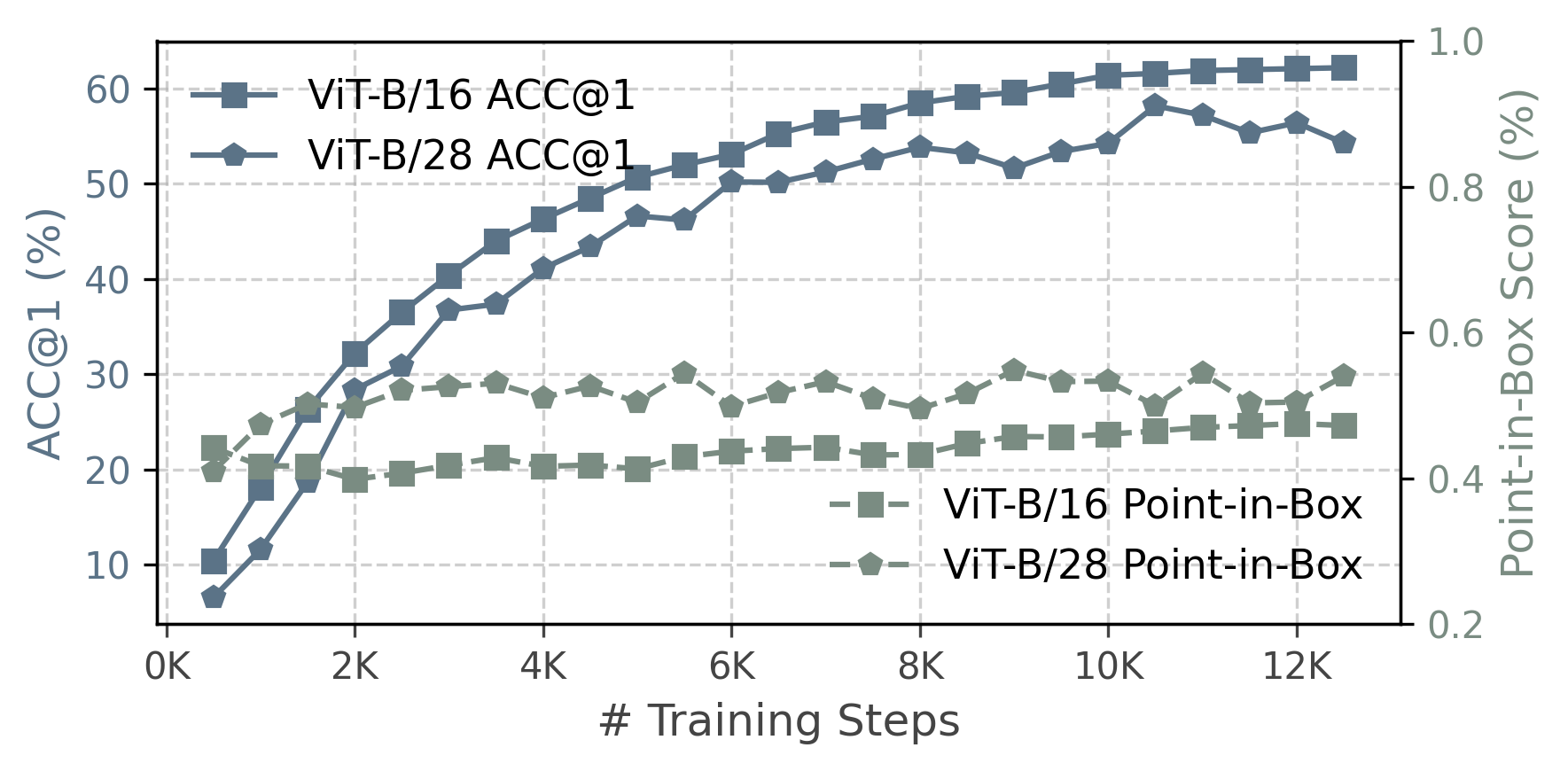}
\vspace{-8pt}
\caption{\textbf{Effect of Coarse-grained semantic supervision.}
Increasing the patch size reduces background tokens by $10\%$.
\emph{Effect:} Point-in-Box rises from $0.44$ to $0.52$, and high-score patches shift toward foreground.
\emph{Trade-off:} classification accuracy decreases, indicating that coarse-grained semantic supervision contributes to artifacts, while naive patch coarsening compromises recognition.}
\label{fig:stagev2}
\end{figure}

\noindent \textbf{\textcolor{black}{Validation Experiment Results.}} 
As shown in Fig.~\ref{fig:stagev2}, Point-in-Box increases from $0.44$ to $0.52$ after enlarging the patch size, which reduces the proportion of background tokens by about $10\%$.
Patch-score maps show that high-score regions shift from background to object areas.
However, top-$1$ accuracy drops from $62\%$ to $55\%$, revealing a trade-off  between classification and localization accuracies.

\subsection{Lazy Behavior from ViT’s Global Dependencies}
\label{sub:window}

\noindent \textbf{\textcolor{black}{Validation Experiment Setting.}}
We further examine whether ViT’s global attention exacerbates the \textit{lazy aggregation} behavior by allowing foreground semantics to be propagated into background regions.
To progressively restrict long-range dependencies, we replace global self-attention with window-based attention~\cite{liu2021swin} at different layers.

\begin{table}[h]
\centering
\small
\begin{adjustbox}{max width=0.99\linewidth}
\begin{tabular}{cccc}
\toprule
\textbf{Replaced Layer} & \textbf{Window Size} & \textbf{Top-1 (IN1K)} & \textbf{Point-in-Box} \\
\midrule
None      & None & 72.3 & 50.1 \\
1, 5, 9, 11 & 4   & 71.7 & 52.1 \\
All       & 4    & 63.9 & 59.8 \\
\bottomrule
\end{tabular}
\end{adjustbox}
\caption{\textbf{Window-attention ablation on ViT-Small.}
Restricting global dependencies raises Point-in-Box but reduces top-1 accuracy.
This suggests that unrestricted global attention amplifies lazy behavior, as coarse-grained semantic supervision allows background tokens to absorb diffused semantics from the foreground.}
\label{app:tab:hyper}
\end{table}

\noindent \textbf{\textcolor{black}{Validation Experiment Results.}}
As shown in Tab.~\ref{app:tab:hyper}, the Point-in-Box score increases as global attention is limited, with the highest value achieved when all layers adopt window attention.
However, accuracy declines correspondingly, implying that while global context benefits classification, it also facilitates semantic diffusion into background patches.

\begin{conclude}
\textit{\textbf{Takeaway:} 
When vision transformers are trained under coarse-grained semantic supervision and equipped with unrestricted global dependencies, 
they tend to pursue the easiest optimization path---diffusing foreground semantics into abundant background tokens. 
This shortcut yields high image-level accuracy but results in representations dominated by background information, showing that optimizing for global classification objectives can compromise patch-level semantic consistency in ViTs.}
\end{conclude}

\section{Method}
\label{m51}

\noindent\textbf{Overview and Rationale.}
To mitigate lazy aggregation, we reformulate \texttt{CLS} token aggregation as a frequency-aware process that distinguishes foreground patches from background ones. 
In natural images, foreground signals have more homogeneous semantic meaning, giving rise to less variations along the channel dimension of a feature map in a deep layer, whereas background often has higher semantic diversity; thus selecting tokens that are stable under low-pass filtering in the channel dimension can potentially anchor \texttt{CLS} tokens to foreground regions.

\subsection{\ours}
\label{sub:fouri}

\noindent\textbf{Stability Score.}
Let $\mathbf{x}_{\mathrm{patch}}\in\mathbb{R}^{N\times D}$ denote the collection of all patch representations generated from the ViT encoder (after dropping \texttt{[CLS]}) and let $\mathbf{g}\in[0,1]^D$ be a normalized vector of Gaussian weights duplicated to all patches:
\begin{equation}
\begin{aligned}
\mathbf{x}_{\mathrm{FFT}}      &= \mathrm{FFT1D}(\mathbf{x}_{\mathrm{patch}}),\\
\mathbf{x}_{\mathrm{LP}}       &= \mathbf{x}_{\mathrm{FFT}} \odot \mathbf{g},\\
\hat{\mathbf{x}}_{\mathrm{patch}}&= \Re\{\mathrm{IFFT1D}(\mathbf{x}_{\mathrm{LP}})\},
\end{aligned} \tag{4} 
\label{eq:lowpass}
\end{equation}
where FFT1D and IFFT1D respectively represent the 1D Fourier transform and the 1D inverse Fourier transform in the channel dimension of every patch, $\odot$ is element-wise multiplication, and $\Re\{\cdot\}$ extracts the real part. The channel-wise stability score compares individual channels of original and low-pass-filtered patch representations:
\begin{equation}
\mathbf{S}_{i,j}=\frac{\hat{\mathbf{x}}_{\mathrm{patch}}[i,j]}{\bigl|\hat{\mathbf{x}}_{\mathrm{patch}}[i,j]-\mathbf{x}_{\mathrm{patch}}[i,j]\bigr|+\varepsilon}, \tag{5} 
\end{equation}
where $i$ is the patch index and $j$ is the channel index.

\noindent\textbf{Channel-wise Top-$K$ Pooling.}
Using channel-wise stability scores, we aggregate patch representations into the \texttt{CLS} token by selecting, for each channel, the $K$ most stable patches (tokens) and averaging them:
\begin{equation}
\mathcal{I}_{K}(j)\;=\;\operatorname{TopK}\!\big(\{\mathbf{S}_{i,j}\}_{i=1}^{N},\,K\big),\qquad j=1,\ldots,D, \tag{6} 
\end{equation}
\begin{equation}
\begin{aligned}
\mathcal{Q}_{\text{CLS}}[j]
&= \operatorname{\!Pool}_{K}\!\big(\mathbf{x}_{\mathrm{patch}}[:,j];\,\mathbf{S}_{:,j}\big)\\
&\triangleq \frac{1}{K}\sum_{i\in \mathcal{I}_{K}(j)} \mathbf{x}_{\mathrm{patch}}[i,j],\qquad j=1,\ldots,D,
\end{aligned} \tag{7} 
\end{equation}
where $\mathcal{I}_{K}(j)$ represents the index set of the $K$ patches with the highest stability scores in the $j$-th channel.


\noindent\textbf{Vote Count.}
We define the vote count of token (patch) $i$ as
\begin{equation}
\label{eq:vote}
v_i \;\triangleq\; \sum_{j=1}^{D} \mathbf{1}\!\bigl\{\, i \in \mathcal{I}_{K}(j) \,\bigr\}, \qquad i=1,\ldots,N, \tag{8} 
\end{equation}
where $\mathbf{1}\{\cdot\}$ denotes the indicator function. A larger $v_i$ indicates a greater importance of patch $i$ among all patches.

\begin{figure}[t]
\scriptsize
\centering

\includegraphics[width=0.48\textwidth]{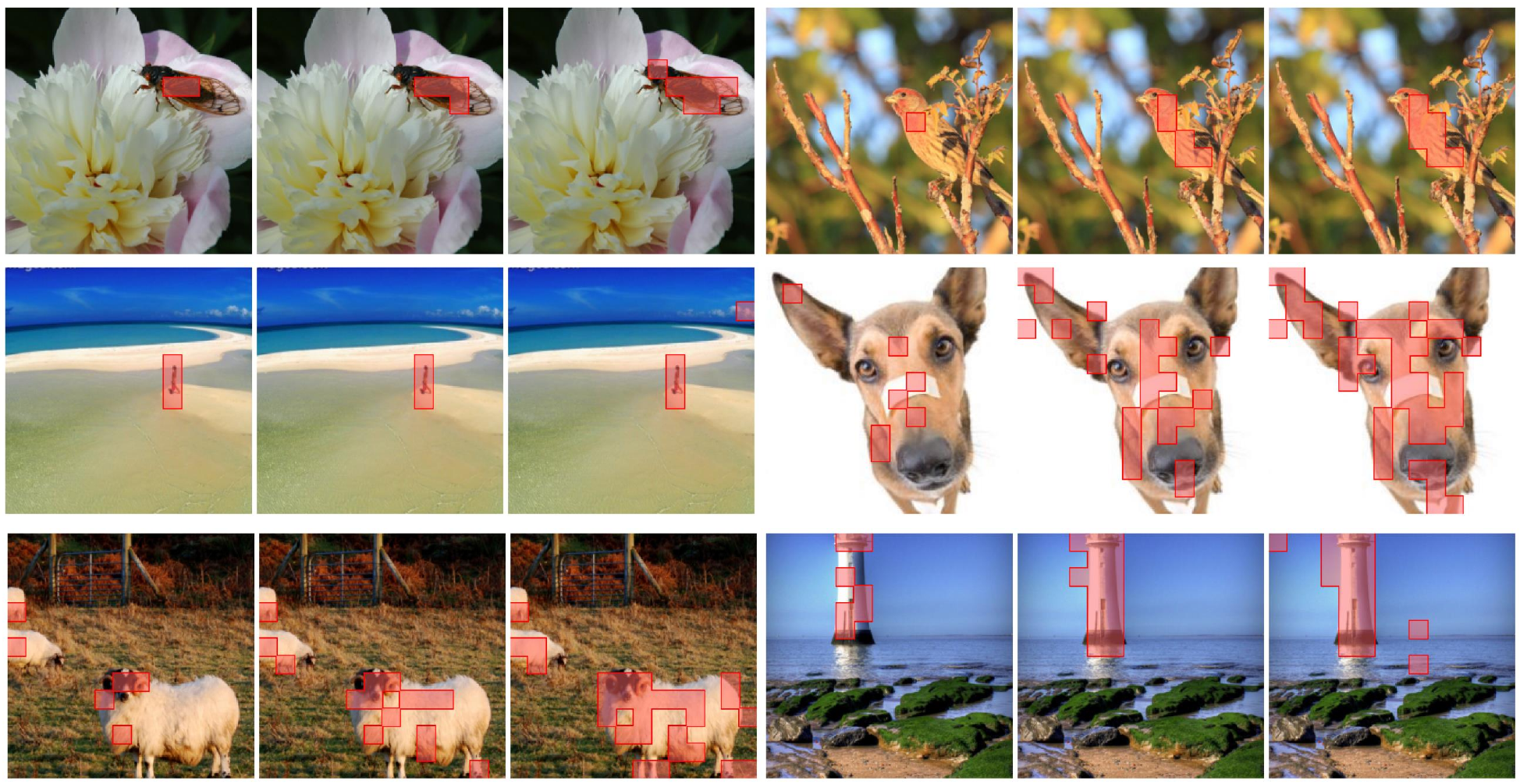}
\caption{\textbf{Where does the \texttt{CLS} token in \ours ``look at"?}
For each image, patches whose vote count exceeds 50\%, 30\%, or 20\% of the largest vote count within the image are visualized in red from left to right, respectively.
After the application of \ours, highly voted patches consistently correspond to foreground regions, 
showing that the \texttt{CLS} token primarily aggregates foreground tokens rather than background ones.}
\label{fig:cls_after_lazystrike}
\vspace{-4pt}
\end{figure}

\paragraph{Where does the \texttt{CLS} token in \ours look at?}
After the application of \ours, the highly voted patches are better aligned with the foreground regions and the number of such patches increases or decreases with the amount of foreground evidence (see Fig.~\ref{fig:cls_after_lazystrike}), indicating that the model has learned to anchor the \texttt{CLS} token to the foreground patches.

\subsection{Transfer to Downstream Tasks}
In this section, we provide further details and explain how each downstream task is conducted.

\noindent Unsupervised Object Discovery. Since \ourmethod guides the \texttt{CLS} token to focus on foreground objects, we can achieve unsupervised object localization using patch scores. \textit{This expansion is independent of the training method—typically a privilege of self-supervised approaches like DINO in earlier works—\textbf{allowing any training objective to accomplish this.}} We construct the mask by applying a threshold defined as the mean score plus one standard deviation. Patches with scores above this threshold are classified as foreground.

\noindent Zero-shot Open-Vocabulary Tasks. Since \ourmethod ensures that the \texttt{CLS} feature aggregates information from the correct patch features, and the \texttt{CLS} feature itself is directly supervised by the learning signal, this effectively leads to an implicit alignment between patch features and the supervision signal. For text-supervised ViTs, we can obtain zero-shot semantic segmentation results by computing the similarity between patch features and arbitrary text features, thereby enabling applications across various open-vocabulary tasks.

\section{Experiment}
\subsection{Experiment Settings}
We first verify the elimination of artifacts in patch score (Sec.~\ref{exp:artifact}) and validate our proposed method on three training methods: fully supervised (Sec.~\ref{sec:full}), text-supervised (Sec.~\ref{sec:weak}), and self-supervised (Sec.~\ref{sec:self}),
and examine multiple downstream tasks for ViT under different supervision, including object discovery~\cite{lost,dinov1}, zero-shot semantic segmentation~\cite{clip,eva,metaclip}, open-vocabulary object detection~\cite{fvlm,shi2024devil}, instance segmentation~\cite{wu2023clipself,shi2023edadet, } and coarse segmentation~\cite{dinov1}. 




\begin{table}[t]
    \centering
    \begin{adjustbox}{max width=0.87\linewidth}
    \begin{tabular}{lcc}
    \toprule
    	Method   & High Norm  & Points-in-Box \\
    	\midrule
        \color{gray}ResNet~\cite{resnet}   & \color{gray}\ding{55} & \color{gray}68.4  \\
         ViT~\cite{vit}   & \ding{51} & 42.7  \\
         \res{\text{ViT}}{+\ourmethod}  & \ding{55} & \tablefirst \res{55.1}{+12.4} \\\midrule
         \color{gray}DINO-ResNet~\cite{dinov1}   & \color{gray}\ding{55} & \color{gray}71.1 \\
        DINO-v1~\cite{dinov1}    & \ding{55} & 44.5\\ 
        \res{\text{DINO-v1}}{+\ourmethod} & \ding{55} & \tablefirst\res{69.7}{+25.2} \\\midrule
        \color{gray}CLIP-ResNet~\cite{clip}  & \color{gray}\ding{55} & \color{gray}53.9\\ 
      CLIP~\cite{clip}   & \ding{51} & 39.8\\ 
              \res{\text{CLIP}}{+\ourmethod}  & \ding{55} & \tablefirst \res{50.1}{+10.3}\\
     \bottomrule
    \end{tabular}
    \end{adjustbox}
    \caption{Evaluation of the \ourmethod in Points-in-Box score.
    }\vspace{-5pt}
    \label{tab:pibv2}
\end{table}

\begin{figure}[t]
\scriptsize
\centering
\includegraphics[width=0.48\textwidth]{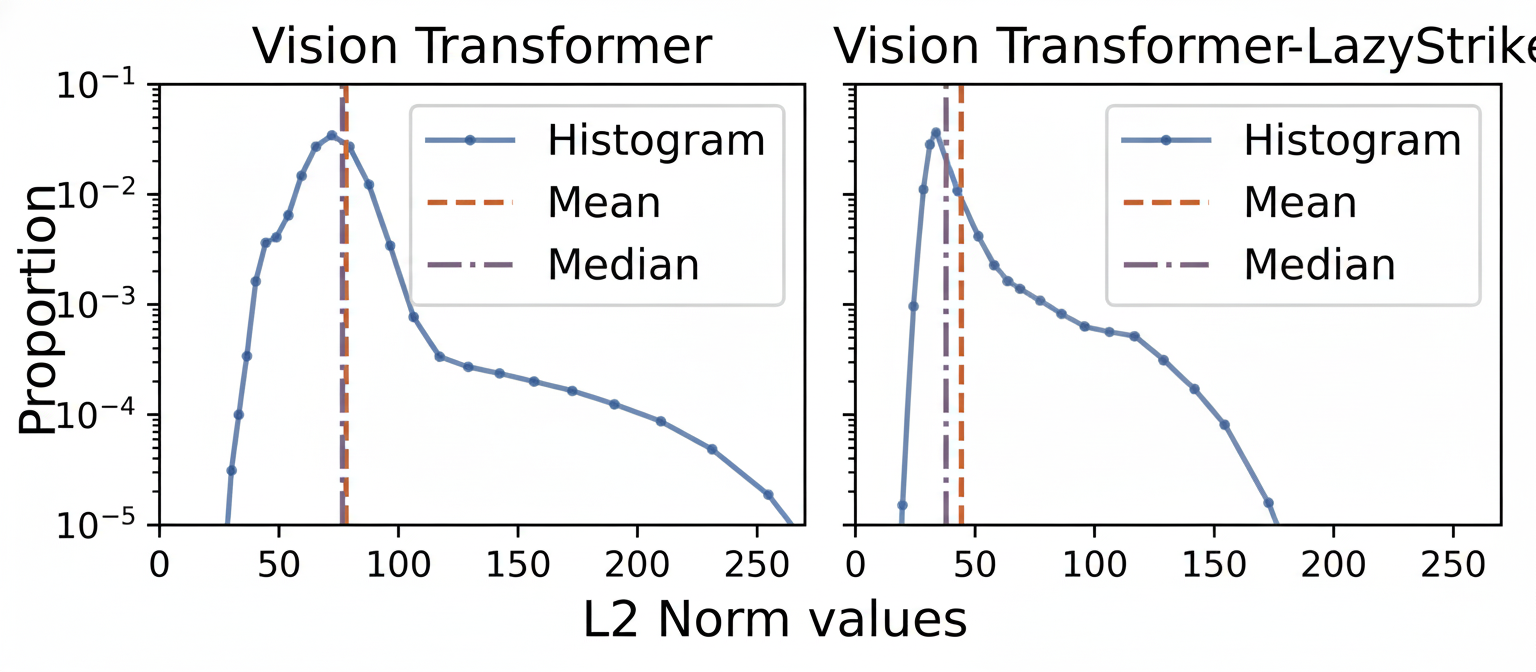}
\caption{Evaluation of the \ours in feature norm. Specifically, the elimination of artifacts also removes the high-norm phenomena~\cite{registers}, highlighting our deeper perspective on addressing artifacts.}\vspace{-5pt}
\label{fig:norm_visv0}
\end{figure}

\begin{table*}[t]
\small
    \centering
    \setlength{\tabcolsep}{6.0pt}
    \begin{tabular}{lc:cccccc}
    \textbf{Model} & Backbone 
     &  \textbf{COCO-Obj.} & \textbf{ADE20K}& \textbf{City.} & \textbf{VOC20} & \textbf{Context59}   &  \textbf{COCO-Stf.}  \\\midrule
    CLIP~\cite{clip}& ViT-B/16 & 8.8 & 3.1   & 6.5 & 49.0 & 11.2  & 7.2 \\
    \res{\text{CLIP}}{+\ourmethod}  & ViT-B/16 &\tablefirst \tablefirst\res{\text{13.3}}{+4.5} & \tablefirst\res{\text{8.3}}{+5.2} & \tablefirst\res{\text{12.1}}{+5.6}& \tablefirst\res{\text{75.0}}{+26.0}   & \tablefirst\res{\text{15.2}}{+4.0} & \tablefirst\res{\text{11.8}}{+4.6} \\
    
    MetaCLIP~\cite{metaclip} & ViT-B/16 & 4.8 & 2.9 & 5.8 & 39.6 & 9.3  &6.2 \\
    \res{\text{MetaCLIP}}{+\ourmethod}  & ViT-B/16 & \tablefirst\res{\text{14.1}}{+9.3} & \tablefirst\res{\text{7.9}}{+5.0} & \tablefirst\res{\text{11.1}}{+5.3}& \tablefirst\res{\text{72.8}}{+33.2}   & \tablefirst\res{\text{15.5}}{+6.2} & \tablefirst\res{\text{12.0}}{+5.8} \\

    EVACLIP~\cite{eva}  & ViT-B/16 & 15.0 & 6.7 & 12.2 & 56.5 & 14.1  & 9.7 \\
    \res{\text{EVACLIP}}{+\ourmethod}  & ViT-B/16 & \tablefirst\res{\textbf{26.2}}{+11.2} &  \tablefirst\res{\textbf{14.8}}{+8.1} &  \tablefirst\res{\textbf{24.5}}{+12.3}& \tablefirst\res{\textbf{79.6}}{+23.1}  & \tablefirst\res{\textbf{24.7}}{+10.6} & \tablefirst\res{\textbf{18.3}}{+8.6} \\
    
    \midrule
    CLIP~\cite{clip}& ViT-L/14 & 3.0 & 1.6 & 2.7 & 17.1 & 5.1  & 3.2 \\
    \res{\text{CLIP}}{+\ourmethod}  & ViT-L/14 &\tablefirst\res{\text{15.0}}{+12.0} & \tablefirst\res{\text{8.4}}{+6.8} & \tablefirst\res{\text{12.3}}{+9.6}&\tablefirst\res{\text{72.4}}{+55.3}   & \tablefirst\res{\text{15.1}}{+10.0} & \tablefirst\res{\text{11.9}}{+8.7}\\

    MetaCLIP~\cite{metaclip} & ViT-L/14 & 5.0 & 3.3 & 6.2 & 25.7 & 8.9  & 6.1\\
    \res{\text{MetaCLIP}}{+\ourmethod}  & ViT-L/14 &
    \tablefirst\res{\text{13.9}}{+8.9} & \tablefirst\res{\text{9.2}}{+5.9} & \tablefirst\res{\text{13.9}}{+7.7}& \tablefirst\res{\text{75.6}}{+49.9}  & \tablefirst\res{\text{16.0}}{+7.1} & \tablefirst\res{\text{12.5}}{+6.4}\\

    EVACLIP~\cite{eva}  & ViT-L/14 & 15.7 & 8.4 & 13.8 & 53.8 &  16.6 & 10.1\\
    \res{\text{EVACLIP}}{+\ourmethod}  & ViT-L/14 &
    \tablefirst\res{\textbf{24.0}}{+8.3} & 
    \tablefirst\res{\textbf{11.3}}{+2.9} &  
    \tablefirst\res{\textbf{17.7}}{+3.9}& 
    \tablefirst\res{\textbf{76.4}}{+22.6}   & 
    \tablefirst\res{\textbf{21.7}}{+5.1}  & 
    \tablefirst\res{\textbf{14.8}}{+4.7}\\
    
    \end{tabular}
    \caption{Evaluation results (mIoU, \%) on six \textbf{semantic segmentation benchmarks}. Our results are marked in \colorbox{gray!10}{gray}. \ourmethod consistently improves semantic segmentation results under text supervision across different type of CLIP~\cite{clip} and model sizes, demonstrating that, after understanding the essence of the problem, a simple approach can uniformly address issues across different models.} 
    \label{tab:semantic}
     
\end{table*}

\begin{table*}[t]
\small
\begin{center}

\setlength{\tabcolsep}{1.7pt}
\begin{tabular}{lc:ccccccccc}
\multirow{2}{*}{\textbf{Method}} & \multirow{2}{*}{Backbone}  & \multicolumn{3}{c}{{\textbf{COCO} Detection}} && \multicolumn{4}{c}{{\textbf{LVIS} Segmentation}} \\
\cline{3-5} \cline{7-10}
& & \multicolumn{1}{c}{AP50$^{\text{box}}$} & \multicolumn{1}{c}{AP50$^{\text{box}}_{\text{base}}$} & \multicolumn{1}{c}{AP50$^{\text{box}}_{\text{novel}}$} 
&&  \multicolumn{1}{c}{AP$^{\text{mask}}$} & \multicolumn{1}{c}{AP$^{\text{mask}}_{\text{freq}}$} & \multicolumn{1}{c}{AP$^{\text{mask}}_{\text{comm}}$} & \multicolumn{1}{c}{AP$^{\text{mask}}_{\text{novel}}$}  \\
\shline

\multicolumn{1}{l}{\bf \textit{\demph{ConvNet based}}} &&&&& \\
\demph{F-VLM}~\cite{fvlm} & \demph{RN50} &    \phantom{1}\demph{39.6} &  \phantom{1}\demph{/} &  \phantom{1}\demph{28.0}  &&  \phantom{1}\demph{24.2} &  \phantom{1}\demph{26.9} &  \phantom{1}\demph{24.0} &  \phantom{1}\demph{18.6} \\



\demph{F-VLM}~\cite{fvlm} & \demph{RN50x64} &  \phantom{1}\demph{/}   &  \phantom{1}\demph{/}&   \phantom{1}\demph{/}  &&  \phantom{1}\demph{34.9} &   \phantom{1}\demph{/}  &  \phantom{1}\demph{/}  &  \phantom{1}\demph{32.8} \\
\hline
\multicolumn{1}{l}{\bf \textit{ViT based}} &&&&&&& \\
   F-ViT~\cite{wu2023clipself} & ViT-B/16 &    \phantom{1}34.9 &  \phantom{1}41.0 &  \phantom{1}17.5  &&  \phantom{1}15.4 &  \phantom{1}20.6 &  \phantom{1}12.3 &  \phantom{1}11.5  \\
 \res{\text{F-ViT}}{+\ourmethod} & ViT-B/16  &  \tablefirst\phantom{1}\res{45.7}{+10.8}	&	\tablefirst\phantom{1}\res{50.1}{+11.1}	&	\tablefirst\phantom{1}\res{33.3}{+15.8} &&  \tablefirst\phantom{1}\res{21.7}{+6.3}	&	 \tablefirst\phantom{1}\res{25.2}{+4.6}	&	 \tablefirst\phantom{1}\res{18.0}{+5.7} &  \tablefirst\phantom{1}\res{22.8}{+11.3} 	\\

 F-ViT~\cite{wu2023clipself} & ViT-L/14 &    \phantom{1}46.0 &  \phantom{1}53.6 &  \phantom{1}24.7  &&  \phantom{1}28.7 &  \phantom{1}31.5 &  \phantom{1}27.9 &  \phantom{1}24.2  \\
 \res{\text{F-ViT}}{+\ourmethod} & ViT-L/14  &  \tablefirst\phantom{1}\res{\textbf{53.2}}{+7.2}	&	\tablefirst\phantom{1}\res{\textbf{68.2}}{+14.6}	&	\tablefirst\phantom{1}\res{\textbf{39.1}}{+14.4} && \tablefirst\phantom{1}\res{\textbf{34.3}}{+5.4}	&	\tablefirst\phantom{1}\res{\textbf{35.1}}{+3.6}	&	\tablefirst\phantom{1}\res{\textbf{34.4}}{+6.6} & \tablefirst\phantom{1}\res{\textbf{32.1}}{+6.6} 	\\
\end{tabular}
\caption{Evaluation results on \textbf{open-vocabulary benchmark}. 
Our results are marked in \colorbox{gray!10}{gray}. \ourmethod consistently enhances performance on open-vocabulary dense tasks, by demonstrating that frozen ViT can achieve comparable performance with ConvNet~\cite{resnet}. }\vspace{-14pt}
\label{tab:ovd}
\end{center}
\end{table*}
\begin{table}[h]
    \centering
    \begin{adjustbox}{max width=1\linewidth}
    \begin{tabular}{llc}
    \toprule
    	Model & Train & mIoU \\
    	\midrule
    	 ViT-B/16  & \textit{Supervised}   &22.3  \\
         \res{\text{ViT-B/16}}{+\ourmethod}  & \tablefirst\textit{Supervised}   & \tablefirst \res{\text{32.8}}{+10.5}\\
         ViT-S/16  & \textit{Supervised}  & 29.5 \\
    	 \res{\text{ViT-S/16}}{+\ourmethod}  & \tablefirst\textit{Supervised}  & \tablefirst \res{\text{41.9}}{+12.4}  \\
         \midrule
         \color{gray}ViT-S/16  & \color{gray}\textit{DINO}   &\color{gray}47.7  \\
         \color{gray}\res{\text{ViT-S/16}}{+\ourmethod}  & \color{gray}\tablefirst\textit{DINO}   & \color{gray}\tablefirst \res{\text{55.1}}{+7.4}\\
     \bottomrule
    \end{tabular}
    \end{adjustbox}
    \caption{\textbf{Coarse segmentation} via patch score. We follow~\cite{white} to conduct coarse segmentation on VOC12. With \ourmethod, ViT under label-supervision also appears emergence of segmentation.} \vspace{-14pt}
    \label{tab:full}
\end{table}

\subsection{Artifact Elimination}
\label{exp:artifact}
\noindent \textbf{Elimination of artifacts in feature norm and patch score.} Tab.~\ref{tab:pibv2} presents the results under different training methods, demonstrating that \ourmethod not only eliminates the high-norm phenomenon but also enhances Point-in-Box score. With \ourmethod applied, ViT’s Point-in-Box score approaches that of ResNet~\cite{resnet}. Fig.~\ref{fig:norm_visv0} provides a detailed analysis of feature norms under fully supervised training~\cite{deit}, revealing that \ourmethod reduces the maximum feature values, thereby mitigating the high-norm phenomenon.

\subsection{Fully-Supervised Comparison}
\label{sec:full}
\noindent \textbf{Emergence of Coarse Segmentation.}
Following \cite{dinov1}, we evaluate emerging properties, a phenomenon only appears in self-supervised training before, on the validation set of VOC12. 
As shown in Tab.~\ref{tab:full}, our method consistently improves emerging properties across different model sizes and training methods. Notably, our approach achieves performance close to DINO in the supervised setting (41.9\% vs. 47.7\%), demonstrating that \ourmethod prompts emerging properties and those are not exclusive to self-supervised.

\begin{figure}[t]
\scriptsize
\centering

\includegraphics[width=0.48\textwidth]{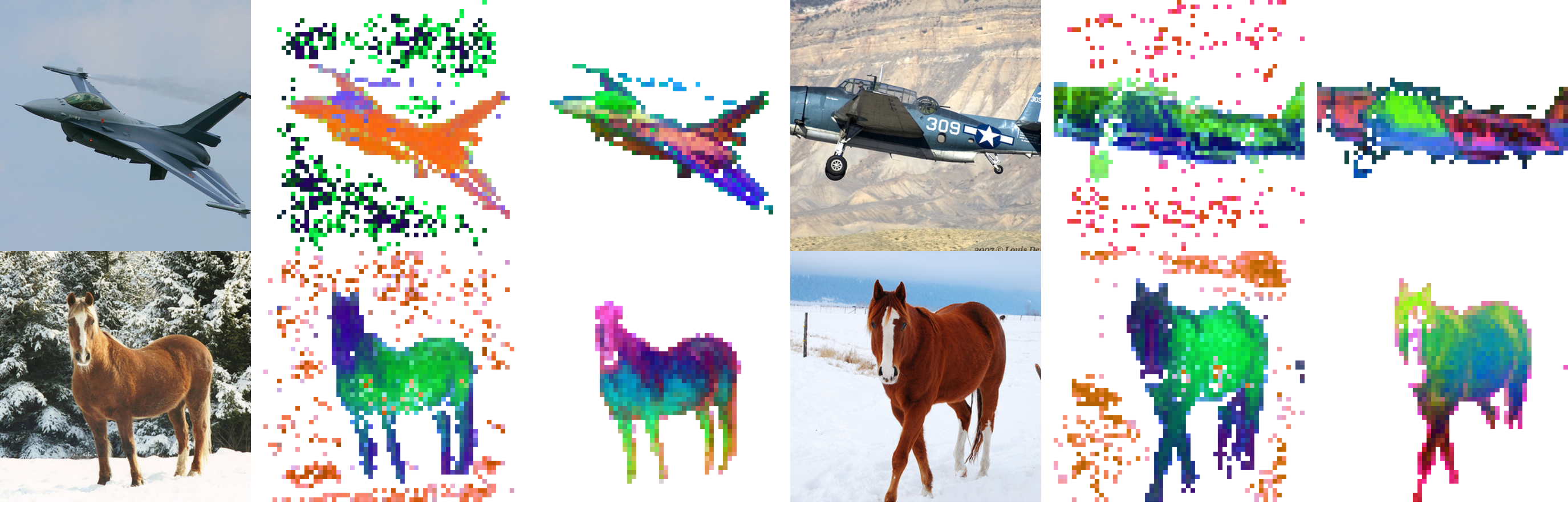}
\put(-193, -5){\textcolor{black}{ViT}~\cite{vit}}
\put(-162, -5){\textcolor{black}{+\textbf{\ourmethod}}}
\put(-70, -5){\textcolor{black}{ViT}~\cite{vit}}
\put(-38, -5){\textcolor{black}{+\textbf{\ourmethod}}}
\caption{\textbf{Visualization of PCA components.} We compute the PCA of the patch features and visualize the first 3 components for the foreground object. With \ourmethod, ViT under label-supervision also distinguish foreground from background and separate object parts, enhancing feature representation.
    }
\label{fig:pca_vis}
\vspace{-8pt}
\end{figure}

\noindent \textbf{Emergence of PCA.} 
As shown in Fig.~\ref{fig:pca_vis}, we compute the PCA of the patch features from \ours and visualize the first three components for the foreground. LazyStrike refines the previously entangled PCA features, \textit{effectively distinguishing and highlighting the salient foreground.}

\subsection{Weakly-Supervised Comparison}
\label{sec:weak}
\noindent \textbf{Zero-shot Semantic Segmentation benchmarks.}
Tab.~\ref{tab:semantic} illustrates our proposed method against several baseline models on six semantic segmentation benchmarks. The improvements achieved by integrating our modifications into these models are highlighted in \textcolor{gain}{blue}.
Our method consistently outperforms the baseline models across all evaluated benchmarks, demonstrating significant gains. For instance, when applied to the CLIP~\cite{clip} model with ViT-B/16 architecture, our method achieves a substantial increase in mIoU on the Pascal (from 11.2\% to 15.2\%), Cityscapes (from 6.5\% to 12.1\%), and VOC (from 49.0\% to 75.0\%). 
When scaled up to the larger ViT-L architecture, our method continues to deliver remarkable results. For the CLIP model, the mIoU on VOC jumps from 17.1\% to an impressive 72.4\%, and on Cityscapes, it increases from 2.7\% to 12.3\%. 
In summary, integrating our method into the baseline models results in significant improvements across all benchmarks, demonstrating its robustness and effectiveness across various CLIP models and models of different sizes.


\noindent \textbf{Open-vocabulary Object Detection and Segmentation benchmarks.} 
As shown in Tab.~\ref{tab:ovd}, We choose F-VLM~\cite{fvlm} and F-ViT~\cite{wu2023clipself} as baselines. Both methods use a \textbf{frozen CLIP~\cite{eva}} as the backbone for object detection and instance segmentation. After obtaining the region of interest, they weigh the semantic scores of the corresponding area to determine the object class scores. The only difference is that F-VLM uses a ConvNet-based backbone, while F-ViT employs a ViT-based backbone. 
For OV-COCO, \ours achieves a gain of 15.8\% and 14.4\% over the baseline on the novel category for ViT-B and ViT-L, respectively. For OV-LVIS, it also improves the baseline by 11.3\% and 6.6\% over the rare category for ViT-B and ViT-L.

\subsection{Self-Supervised Comparison}
\label{sec:self}
\begin{table}
\centering
\small
\begin{adjustbox}{max width=0.95\linewidth}
\begin{tabular}{lcccc} 
\toprule
Method    &FPS   & VOC07 & VOC12 & COCO \\ \midrule
SS~\cite{uijlings2013selective}  & -& 18.8 & 20.9 & 16.0 \\
EdgeBoxes~\cite{zitnick2014edge}  & -    & 31.1 & 31.6 & 28.8 \\ \midrule
DINO-seg~\cite{dinov1}  & 29.4 & 45.8 & 46.2 & 42.1   \\
LOST~\cite{lost}  & 29.4 & 61.9 & 64.0 & 50.7   \\


\res{\text{DINO}}{+\ourmethod} & \tablefirst\textbf{55.9}& \tablefirst\textbf{64.4} & \tablefirst\textbf{67.6} & \tablefirst\textbf{51.6}\\ 
\bottomrule
\end{tabular}
\end{adjustbox}
\caption{\textbf{Object discovery CorLoc}. All models adopt ViT-S. Previous best-performing methods relied on eigenvector computations, whereas \ourmethod avoids such heavy computational demands. } \vspace{-4pt}
\label{tab:corloc_lost}
\end{table}

\noindent \textbf{Unsupervised Object Discovery.}
We adopt DINO-seg~\cite{dinov1} and LOST~\cite{lost} as baselines for comparison, both utilizing ViT-S~\cite{vit} as the backbone for object discovery tasks. The comparisons are illustrated in Tab.~\ref{tab:corloc_lost}. \ours exhibits significant performance improvements.
Specifically, 
our model achieves the highest CorLoc scores across all datasets, surpassing both DINO-seg and LOST models. Notably, our model attains a CorLoc score of 64.4\% on VOC 2007, 67.6\% on VOC 2012, and 51.6\% on COCO, representing improvements of 2.7\%, 3.6\%, and 0.9\% points, respectively, over the best-performing LOST model.
Moreover, our method demonstrates a remarkable throughput of 55.9 images per second.
This indicates that our model achieves superior object discovery performance and operates more efficiently, making it highly suitable for practical applications.


\begin{table}[t]
    \centering\small
    \begin{adjustbox}{max width=0.9\linewidth}
    \begin{tabular}{lccc}
        \toprule
        Method  & IN1K~\cite{deng2009imagenet} & VOC~\cite{voc} & COCO~\cite{voc}\\
        \midrule
        Attention-Pool & 55.8 & 10.7 & 3.3\\
        Max-Pool & 53.1 & 71.9 & 12.2\\
        \midrule
        w/ \ourmethod   \\
        \; \; K = 1 & 53.5 & 72.7 & 13.5 \\
        \; \; K = 49 & \tablefirst55.8 & \tablefirst75.8 & \tablefirst\textbf{18.5}\\
        \; \; K = 98 & \textbf{56.2} & \textbf{75.9} & 18.0\\
        \; \; K = 196 (Full) & 55.3 & 13.5 & 4.8\\
        \bottomrule
    \end{tabular}
    \end{adjustbox}
    \caption{\textbf{Ablation study} on text-supervised ViT~\cite{openclip}. 
    We report ImageNet classification and downstream semantic segmentation results, where \ourmethod significantly addresses the artifact issue and even leads to an improvement in classification performance.
    }
    \label{tab:ablations}
\end{table}

\subsection{Ablation study}

\begin{table}[t]
    \centering\small
    \begin{adjustbox}{max width=0.9\linewidth}
    \begin{tabular}{lccc}
        \toprule
        Method  & IN1K~\cite{deng2009imagenet} & VOC07~\cite{voc} & VOC12~\cite{voc}\\
        \midrule
        Attention-Pool & 59.1 & 14.1 & 28.7\\
        Mean-Pool & 64.3 & 15.3 & 29.6\\
        \midrule
        w/ \ourmethod   \\
        \; \; K = 1 & 64.6 & 30.4 & 35.6 \\
        \; \; K = 7 & \tablefirst64.8 & \tablefirst 32.1 & \tablefirst 37.6\\
        \; \; K = 49 (Full) & \textbf{64.9} & 15.8 & 30.3\\
        \bottomrule
    \end{tabular}
    \end{adjustbox}
    \caption{\textbf{Ablation study }on label-supervised ViT~\cite{deit}. We report ImageNet classification performance and downstream object location results, where \ourmethod significantly addresses artifacts.
    } \vspace{-8pt}
    \label{tab:ablations2}
\end{table}

\noindent \textbf{Other method to alleviate artifacts.} 
In Tab.~\ref{tab:ablations}, we also report results with Maxpool, which naturally reduces background activations and serves as a strong reference for assessing artifact mitigation. 
While Maxpool brings moderate improvement, our approach achieves substantially higher performance across both classification and segmentation tasks, 
demonstrating that the improvement stems from more effective semantic aggregation rather than a pooling-induced side effect.

\noindent \textbf{Number of cutted tokens.} 
In Tab.~\ref{tab:ablations}, we examine the impact of Top-$K$ by training OpenCLIP~\cite{openclip} ViT-B/16 with different number of $K$. Performance improves significantly with \ourmethod, peaking when half of the tokens are selected. Tab.~\ref{tab:ablations2} shows further ablation studies on label-supervised ViT-B/32, with pretraining on ImageNet-1k and classification performance and CorLoc results.

\section{Conclusion}

We reveal that Vision Transformers often adopt a \emph{lazy aggregation} behavior—%
relying on numerous background patches to encode global semantics due to their overwhelming dominance over foreground regions. 
To counter this, we propose \textbf{LaSt-ViT}, a frequency-guided selective aggregation that focuses the \texttt{CLS} token on stable, foreground-relevant features. 
Our method effectively eliminates artifacts across various supervision types and achieves consistent improvements on 12 benchmarks, providing a clearer understanding of ViT’s internal behavior and a solid baseline for future research.

\section{Acknowledgments}

This work is supported in part by the Hong Kong Research Grants Council under Collaborative Research Fund (Project No. HKUC7004-22G) and the National Natural Science Foundation of
China under Grant No.62576365.

{
    \small
    \bibliographystyle{ieeenat_fullname}
    \bibliography{main}
}


\end{document}